\documentclass[10pt,twocolumn,letterpaper]{article}

\usepackage[pagenumbers]{cvpr} %

\newcommand{\method}{DFM}

\usepackage{amsmath}
\usepackage{algorithm}
\usepackage{algpseudocode}
\usepackage{cvpr}

\usepackage{blindtext}%
\usepackage{graphicx}
\usepackage{caption}

\usepackage{breakcites} %
\usepackage{microtype} %

\usepackage{amsmath,amsfonts,bm,amssymb,mathtools,amsthm}
\usepackage{color,xcolor,xspace}
\usepackage{booktabs}
\usepackage{thm-restate}
\usepackage{bbding}
\usepackage{pifont}
\usepackage{wasysym}

\def\eqref#1{Eq.(\ref{#1})}

\def\1{\bm{1}}

\def\rvv{{\mathbf{v}}}

\def\rvx{{\mathbf{x}}}

\DeclareMathAlphabet{\mathsfit}{\encodingdefault}{\sfdefault}{m}{sl}
\SetMathAlphabet{\mathsfit}{bold}{\encodingdefault}{\sfdefault}{bx}{n}

\def\rvv{{\mathbf{v}}}

\def\rvx{{\mathbf{x}}}

\usepackage{duckuments}
\definecolor{cvprblue}{rgb}{0.21,0.49,0.74}
\usepackage[pagebackref,breaklinks,colorlinks,allcolors=cvprblue]{hyperref}

\def\paperID{6669} %
\def\confName{CVPR}
\def\confYear{2025}

\title{Decentralized Diffusion Models} %

\author{David McAllister$^{1}$\thanks{Work partially done when author was an intern at Luma AI} \quad Matthew Tancik$^2$ \quad Jiaming Song$^2$ \quad Angjoo Kanazawa$^1$\thanks{Author is an advisor to Luma AI} \\[2ex]
\centering
$^1$University of California, Berkeley \hspace{3em} $^2$Luma AI
}

\begin{document}
\maketitle

\begin{abstract}

Large-scale AI model training divides work across thousands of GPUs, then synchronizes gradients across them at each step. This incurs a significant network burden that only centralized, monolithic clusters can support, driving up infrastructure costs and straining power systems. We propose Decentralized Diffusion Models, a scalable framework for distributing diffusion model training across independent clusters or datacenters by eliminating the dependence on a centralized, high-bandwidth networking fabric.
Our method trains a set of expert diffusion models over partitions of the dataset, each in full isolation from one another. At inference time, the experts ensemble through a lightweight router. We show that the ensemble collectively optimizes the same objective as a single model trained over the whole dataset.
This means we can divide the training burden among a number of ``compute islands,'' lowering infrastructure costs and improving resilience to localized GPU failures. Decentralized diffusion models empower researchers to take advantage of smaller, more cost-effective and more readily available compute like on-demand GPU nodes rather than central integrated systems. We conduct extensive experiments on ImageNet and LAION Aesthetics, showing that decentralized diffusion models FLOP-for-FLOP outperform standard diffusion models. We finally scale our approach to 24 billion parameters, demonstrating that high-quality diffusion models can now be trained with just eight individual GPU nodes in less than a week.%

\end{abstract}

\begin{figure}[tp]
\begin{center}
    \centering
    \captionsetup{type=figure}
    \begin{center}
    \centering
    \makebox[\linewidth]{\includegraphics[width=1.1\linewidth]{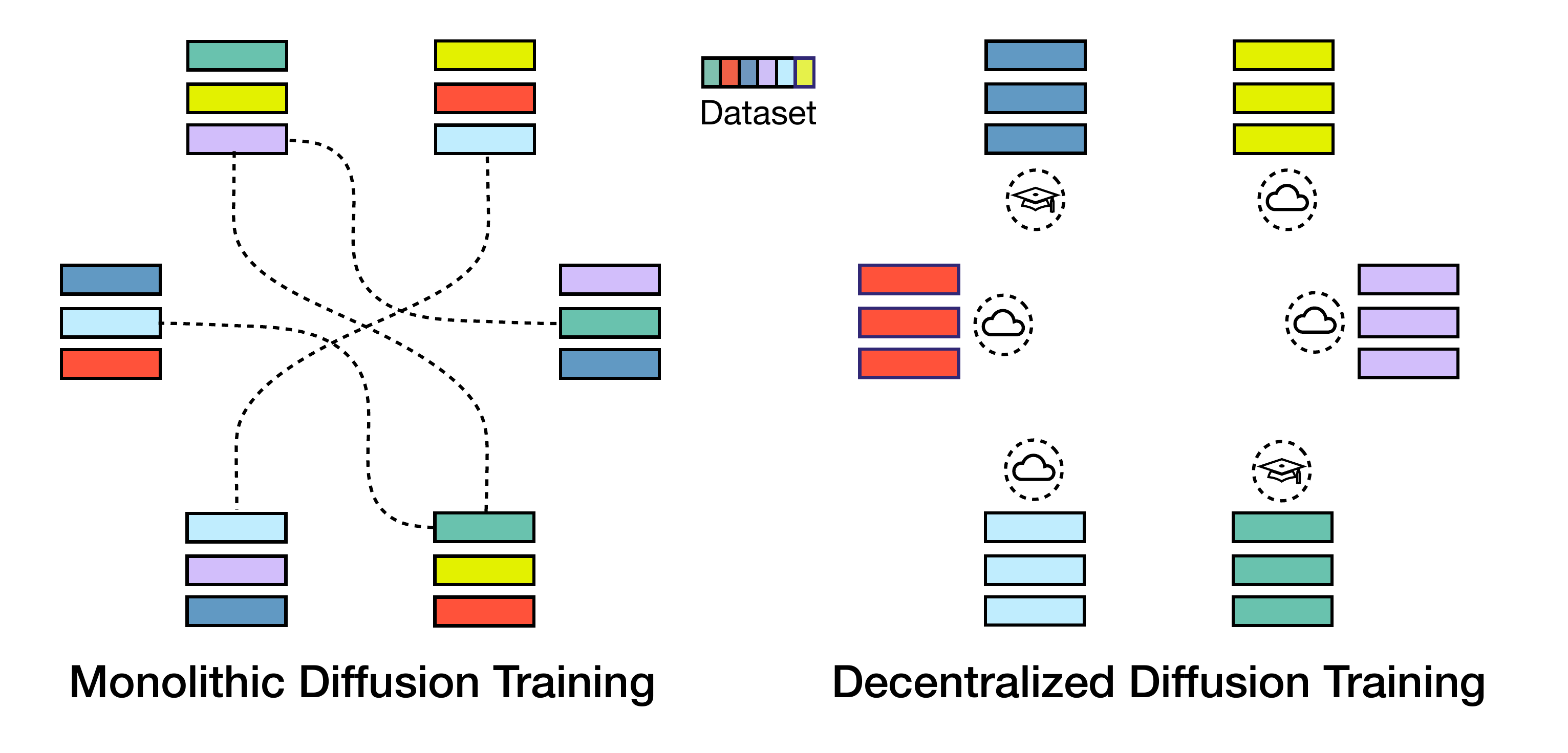}}
    \end{center}
    \vspace{-0.5em}
    \captionof{figure}{\textbf{Decentralized Diffusion Models (DDM).}
    Left: Existing diffusion models (monolithic) require synchronized, centralized training across thousands of GPUs, making high-quality training systems expensive and inaccessible. Right: DDM divides a diffusion model into an ensemble of expert models, each trained on its own data cluster in complete isolation. This ensemble collectively optimizes the same diffusion objective as a single model trained on all the data. This enables flexible training across diverse cloud or academic compute facilities. At inference, the ensemble delivers improved performance at the same FLOP-cost, making high-quality diffusion model training more efficient and accessible. See Figure~\ref{fig:full-page} for large-scale DDM model samples.
    }
    \vspace{-2em}
\end{center}
\end{figure}

\begin{figure*}[t]
    \centering
    \includegraphics[width=\linewidth]{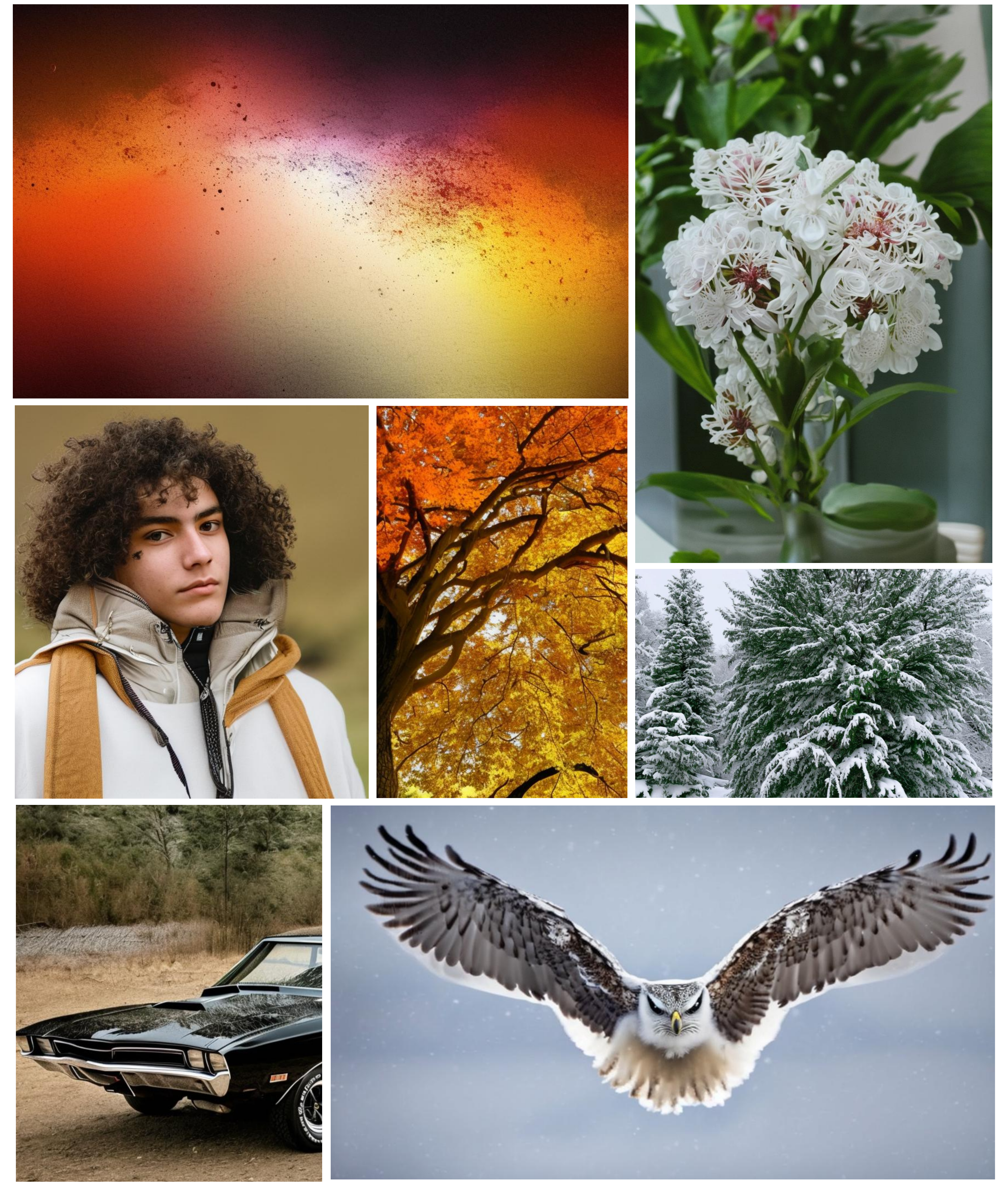}
    \caption{\textbf{Decentralized diffusion models train on readily-available hardware and generate high quality, diverse images.} We present selected samples from our 8x3B parameter model.}
    \label{fig:full-page}
\end{figure*}

\section{Introduction}
\label{sec:intro}

Diffusion models achieve breakthrough results in image generation \cite{rombach2022highresolutionimagesynthesislatent, saharia2022photorealistictexttoimagediffusionmodels}, video modeling \cite{videoworldsimulators2024}, and robotic control through diffusion policies \cite{chi2024diffusionpolicyvisuomotorpolicy}. However, these models continue to demand greater and greater training compute. Even early successes in image diffusion underscored the immense computational requirements, with Stable Diffusion 1.5's training consuming over 6,000 A100 GPU days \cite{podell2023sdxlimprovinglatentdiffusion, chen2023pixartalphafasttrainingdiffusion}. Video diffusion models demand a step-function increase in these resources. For instance, Meta's Movie Gen trains on up to 6,114 H100 GPUs~\cite{polyak2024moviegencastmedia}---nearly GPT-3 scale~\cite{brown2020languagemodelsfewshotlearners}. As these models continue to grow in size and capability, they encounter significant systems-level challenges that mirror those faced in LLM training. High-bandwidth interconnect becomes a critical bottleneck, storage systems must handle massive streaming datasets and persistent hardware failures can crash or silently slow the training process \cite{an2024fireflyeraihpccosteffectivesoftwarehardware, dubey2024llama3herdmodels}. The result is a fragile integrated system where performance and reliability depend on complex interactions between arrays of accelerators and networking hardware. This is challenging and costly for large industry labs---untenable for academics.

We present Decentralized Diffusion Models, a scalable method for distributing the modeling burdening of diffusion across independent expert models, each trained on its own “compute island” with no cross-communication. This enables training with scattered resources, leveraging cost-effective cloud compute or combining resources across multiple clusters. Decentralization is also helpful for training ``foundation''-scale models~\cite{bommasani2022opportunitiesrisksfoundationmodels}, where it is increasingly difficult to build datacenters with enough power capacity for the scale of modern training runs~\cite{Morey_2024}. Independent training provides additional benefits, such as the ability to execute across heterogeneous hardware, making it possible to reuse existing training clusters alongside newer infrastructure.

Decentralized diffusion models employ a new training objective, Decentralized Flow Matching (\method{}), which partitions the training data into K groups and trains a dedicated diffusion model over each. We refer to these specialized models as experts, inspired by the Mixture of Experts (MoE) literature~\cite{fedus2022switchtransformersscalingtrillion}.
An independently trained router model orchestrates these experts, determining which are most appropriate at test-time. We show that this ensemble of expert models collectively optimizes the same global objective as a single model trained on the entire dataset.

During each inference step, the router predicts each expert's relevance to the input noise and condition. The expert predictions are then linearly combined, weighed by their associated relevance scores. Experts learn to specialize, so many are irrelevant to a given input, and it is more efficient to use a subset of them. If not all experts are selected, it serves as a sparse model, leveraging selective computation similar to MoE. Sparse inference is compute-efficient but remains memory-intensive.  We demonstrate that we can distill experts into a single dense model, achieving convenient inference while preserving sample quality. %

We evaluate our approach on ImageNet 
and a filtered subset of LAION-Aesthetics, where we compare DDMs against existing monolithic diffusion model training. We systematically vary the number of experts to analyze its impact on downstream generation tasks, where we find that DDMs with eight experts consistently achieve the best performance, even outperforming a single model trained with the same compute. This is made possible by the additional parametrization provided by multiple specialized experts to outperform a generalist model.  Finally, we demonstrate the scaling ability of DDMs by training a large decentralized model with eight 3-billion parameter experts, resulting in high definition image generation, as shown in Figure~\ref{fig:full-page}. 

Decentralized Diffusion Models also integrate easily into existing diffusion training environments. In practice, DDMs involve clustering a dataset then training a standard diffusion model on each cluster. This means nearly everything from existing diffusion infrastructure can be reused---training code, dataloading, systems optimizations, noise schedules and architectures.
See our \href{https://decentralizeddiffusion.github.io/}{blog post} for a visual walkthrough and pseudocode.

\section{Related Works}
\label{sec:related}

\paragraph{Accelerating Diffusion Models}
Flow matching ~\cite{lipman2023flowmatchinggenerativemodeling, liu2022flowstraightfastlearning} generalizes diffusion models~\cite{sohldickstein2015Diffusion,song2019generative,ho2020denoising,song2021scorebasedgenerativemodelingstochastic} and makes tractable large-scale training of continuous normalizing flows (CNFs) \cite{chenNODE}. %
Diffusion and flow matching models produce state of the art generations in high dimensional continuous spaces but are typically expensive to train to high quality. Recent works propose token masking \cite{sehwag2024stretchingdollardiffusiontraining} and aligning internal states with pretrained image representation models \cite{yu2024representationalignmentgenerationtraining} to accelerate the model learning process. PixArt-$\mathbf{\alpha}$~\cite{chen2023pixartalphafasttrainingdiffusion} shows that curated datasets with synthetic captions reduce training costs. We explore a complementary direction that decentralizes the training of the model across multiple GPU clusters without cross-communication, thereby increasing the robustness and efficiency of training. %

Existing approaches all use data parallel training, which distributes batches across GPUs and synchronizes gradients to produce larger effective batch sizes. Our work introduces an orthogonal form of parallelism that partitions the model's training across isolated compute centers with no gradient synchronization. Our approach is complementary with existing techniques---in fact, we employ Fully Sharded Data Parallel (FSDP)~\cite{zhao2023pytorchfsdpexperiencesscaling} training within each cluster while maintaining isolation between them in our experiments.

\paragraph{Mixture of Experts} Mixture of Experts (MoE) is a popular and powerful approach to increase model capacity without a proportional increase in computational cost. This is achieved through sparse parameter activation: MoE replaces each dense feed forward network (FFN) layer of a transformer with a lightweight router that selects $k$ of $N$ learned FFN layers per token. The approach gained prominence with Switch Transformers~\cite{fedus2022switchtransformersscalingtrillion}, followed by works that refined MoE via various routing strategies~\cite{zhou2022mixtureofexpertsexpertchoicerouting} and systems improvements~\cite{gale2022megablocksefficientsparsetraining}.

These advances have led to notable successes in language modeling, exemplified by models like Mixtral~\cite{jiang2024mixtralexperts} and DeepSeek-V3~\cite{deepseekai2024deepseekv3technicalreport}, which achieve strong performance while maintaining computational efficiency. While our work focuses on decentralized training, we draw inspiration from MoE techniques at test-time, adapting the principle of increasing parametrization through sparsely-activated experts to the diffusion domain. Both MoE and DDMs increase total model capacity without increasing computational costs, but we achieve this by routing inputs to different data experts rather than routing tokens to different FFNs.

MoE architectures introduce significant systems challenges, particularly the problems of expert load balancing, managing capacity factors, and the challenge of holding many times more parameters in memory during training and inference. Conversely, DDMs alleviate systems constraints by training experts and routers individually, extending diffusion training to readily available hardware configurations.

\begin{figure*}[t]
    \centering
    \includegraphics[width=\linewidth]{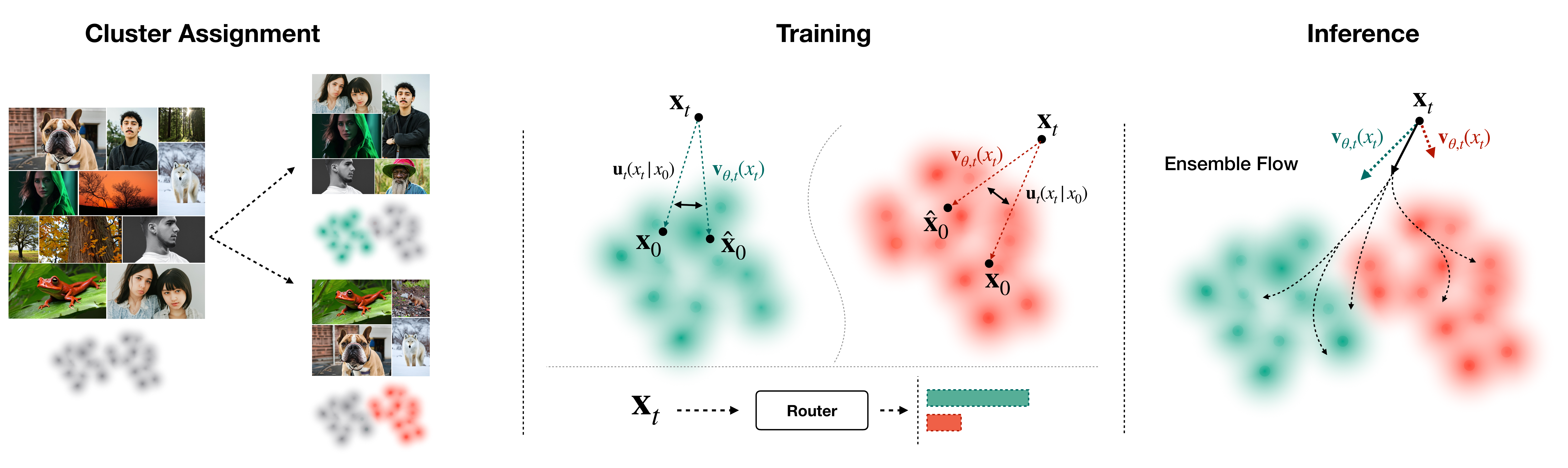}
    \caption{\textbf{Decentralized Diffusion Model (DDM) Training Overview.}
    DDMs follow a three-step training process. We first cluster the dataset using off-the-shelf representation extraction models. We train a diffusion model over each of these clusters and a router that associates any input $x_t$ with its most likely clusters. 
    At test-time, given a noisy sample, each expert (in red and green) predict their own flows, which combine linearly via the weights predicted by the router. The combined flow samples the entire distribution and is illustrated on the right.
}
    \vspace{-0.4cm}
    \label{fig:flow_ensemble}
\end{figure*}

\paragraph{Low Communication Learning} Federated learning ~\cite{mcmahan2023communicationefficientlearningdeepnetworks} establishes foundational techniques for communication-constrained training, originally motivated by data privacy concerns. Follow-up works including Federated Averaging ~\cite{sun2021decentralizedfederatedaveraging}, Adaptive Federated Optimization ~\cite{reddi2021adaptivefederatedoptimization} and DiLoCo ~\cite{douillard2024dilocodistributedlowcommunicationtraining}, adapt these ideas to address large-scale training challenges. DiLoCo, uses an inner and outer optimization loop to balance local training with periodic global synchronization. This parallels developments in large model training, where methods like Branch-Train-Merge ~\cite{li2022branchtrainmergeembarrassinglyparalleltraining} and Diffusion Soup ~\cite{biggs2024diffusionsoupmodelmerging, wortsman2022modelsoupsaveragingweights} have explored training and combining data-specialist models. While our approach shares technical insights with these methods, we focus on leveraging distributed computation for robustness and flexibility rather than data sovereignty or training on mobile devices. In fact, our method should complement these approaches for further decentralization, which we hope to see in future work.

\section{Decentralized Flow Matching}
\label{sec:method}

We introduce Decentralized Flow Matching (\method{}), the training objective for DDMs, to distribute the diffusion modeling burden across an ensemble of expert denoisers, each trained individually without cross-communication. We show that the flow matching objective divides naturally into this distributed arrangement. Since diffusion models and rectified flows can be seen as special cases of flow matching, \method{} applies directly across these popular algorithms. This generalizes a previous derivation for privacy-preserving diffusion models~\cite{golatkar2024trainingdataprotectioncompositional}. Specifically, we train each expert over a pre-determined subset of the data distribution. Each expert's predicted flow is combined at test-time using a router trained with a simple classification objective. The resulting ensemble samples the entire data distribution.

\subsection{Preliminary: Flow Matching Objective}

Flow matching~\cite{lipman2023flowmatchinggenerativemodeling} defines a fixed forward corruption process and a learned reverse denoising process. We index this process with timesteps $t$ interpolating the data distribution at $t=0$ and the per-pixel Gaussian distribution at $t=1$. Concretely, 
$x_t=\alpha_t x_0 + \sigma_t \epsilon,$
\noindent where the scaling coefficients $\alpha_t$ and $\sigma_t$ are chosen ahead of time, typically to preserve variance across timesteps~\cite{song2021scorebasedgenerativemodelingstochastic}. The learned model, $v_{\theta,t}(x_t)$, reverses this corruption process to transport from the noise distribution to the data distribution. These transport paths collectively represent the marginal flow, $u_t(x_t)$, which is a vector field at each timestep that can be regressed in a model to interpolate the data distribution or computed analytically over a dataset, which will reproduce training samples.

Flow transports a latent variable, $x_t$%
, to the data distribution through a multi-step sampling process. At each step, the model predicts the flow from $x_t$ toward a weighted average of the data distribution. In its continuous definition, we express this as an integral:

\begin{align}
    u_t(x_t)=\int_{x_0} u_t(x_t|x_0) \frac{p_t(x_t|x_0)q(x_0)}{p_t(x_t)}dx_0
    \label{eq:marginal_flow_integral}
\end{align}

\noindent Where $u_t(x_t)$ is the marginal flow, $x_t$ is our noisy latent, $x_0$ is a data sample, $p_t(x_t|x_0)$ is Gaussian and $u_t(x_t|x_0)$ is the conditional flow from $x_0$ to $x_t$. Flow matching regresses this analytical form of $u_t$ through a parametric model. Since we train over a discrete dataset, the integral becomes a summation,
\begin{align}
    u_t(x_t)=\frac{1}{p_t(x_t)}\sum_{x_0}u_t(x_t|x_0)p_t(x_t|x_0)q(x_0).
    \label{eq:marginal_flow_sum}
\end{align}

\subsection{Decentralized Flow Matching Objective}

\noindent \method{} decomposes the marginal flow into a series of expert flows. We partition the data into  $K$  disjoint clusters  $\{S_1, S_2, \ldots, S_K\}$, and each expert trains on an assigned subset ($x_0\in S_i$). This is a natural choice, since empirical results suggest that image data lies on a disjoint union of manifolds \cite{brown2023verifyingunionmanifoldshypothesis, kamkari2024geometricviewdatacomplexity} and this may be one reason diffusion models are effective in the image domain \cite{wang2024diffusionmodelslearnlowdimensional}. Splitting across these subsets yields marginal flow of the form:
\begin{align}
    u_t(x_t)=\sum_{k=1}^K \frac{1}{p_t(x_t)} \sum_{x_0 \in S_k}u_t(x_t|x_0)p_t(x_t|x_0)q(x_0).
    \label{eq:dfm1}
\end{align}

\noindent Finally, we find that marginal flow can be written as a linear combination of expert flows,
\begin{align}
    \small
    u_t(x_t)=\underbrace{\sum_{k=1}^K \frac{p_{t,S_k}(x_t)}{p_t(x_t)}}_{\text{Router}} \underbrace{\sum_{x_0 \in S_k}\frac{u_t(x_t|x_0)p_t(x_t|x_0)q(x_0)}{p_{t,S_k}(x_t)}}_{\text{Data-Expert Flow}},
    \label{eq:dfm2}
\end{align}
where the outer sum results in a categorical probability distribution over all experts that we refer to as a `router', and the inner sum is the marginal flow over each subset that we refer to as a data-expert flow. Please see the supplement for an alternative derivation based on score matching.

This division is highly convenient for large-scale training. Instead of one monolith training run, we can train a number of independent expert denoisers. We train each with the standard flow-matching objective and zero cross-model communication. Separately, we train a router that predicts the probability that $x_t$ is drawn from each data subset, which can be formulated as the classification task we describe below. Similar to MoE, this learned router delegates computation to different parameters during inference. A key difference is that our router can be explicitly trained in isolation, instead of being trained end-to-end with gradients flowing through the entire model. %
At inference time, the experts combine in an ensemble that, as we show above, collectively optimize the same global objective as a monolithic training.

\subsection{Router Training}

The router predicts the probability of a given $x_t$ being drawn from each of the data partitions.
\begin{align}
    p(k|x_t,t)=\frac{p_{t,S_k}(x_t)}{p_t(x_t)}, \quad k \in [K]
\end{align}

\noindent To learn this, we sample input $x_t$ then supervise with a cross-entropy loss over the cluster labels associated to each data sample. This loss is minimized when the model, $r_\theta(x_t,t)$, exactly predicts $p(k|x_t,t)$. See Algorithm \ref{alg:router} for details. Note that router trains independently of the denoisers.

We parameterize $r_\theta(x_t,t)$ with a small diffusion transformer (DiT) \cite{peebles2023scalablediffusionmodelstransformers} and use the same conditioning mechanisms as a DiT denoiser. We append a learned classification token \cite{devlin2019bertpretrainingdeepbidirectional} that we decode to cluster logits through a linear head following usual conventions.

\begin{algorithm}

\caption{Flow Router Training}
\begin{algorithmic}[1]
\Require Training data $\{x_0,k\}$, schedule $\{\alpha_t,\sigma_t\}_{t=1}^T$
\While{not converged}
    \State $x_0, k \sim \mathcal{D}$
    \State $t \sim \{1,\ldots,T\}$
    \State $\epsilon \sim \mathcal{N}(0, \mathbf{I})$
    \State $x_t = \alpha_t x_0 + \sigma_t \epsilon$
    \State $\mathbf{z} = r_\theta(x_t, t) \in \mathbb{R}^{|K|}$
    \State Update $\theta$ using $\nabla_\theta\mathcal{L}_{\text{CE}}(\mathbf{z}, \text{OneHot}(k))$
\EndWhile
\end{algorithmic}
\label{alg:router}
\end{algorithm}

\subsection{Expert Training}

The \method{} objective divides the training task into learning a series of expert denoisers. Conveniently, these experts employ the same objective as standard flow matching:
\begin{align}
    \mathcal{L}_{\text{CFM}}(\theta) = \mathbb{E}_{t,q(x_0),p_t(x_t|x_0)} \left\lVert \mathbf{v}_{\theta,t}(x_t) - \mathbf{u}_t(x_t|x_0) \right\rVert^2.
\end{align}

\noindent We parameterize each expert as its own DiT, though it is important to note that our method is architecture-agnostic. We can even assign each denoiser a different architecture as long as it optimizes the above objective.

\subsection{Inference Strategy}

At test-time, we merge the predictions of our experts following the weights predicted by the router:
\begin{align}
    u_t(x_t)=\sum_{k=1}^K \underbrace{r_\theta(x_t,t)}_{\text{Router}} \underbrace{v_{\theta,t}(x_t)}_{\text{Expert}}.
\end{align}

\noindent While the full ensemble is necessary to exactly match the global flow matching objective, it is much less expensive to use a subset of experts or even only one expert for each test-time prediction. The model's FLOP cost scales linearly with the number of active experts in each forward pass. This introduces a trade-off between theoretical correctness and computational efficiency. We can interpret this selection as instantiating model sparsity, where the model activates only a subset of its parameters in each forward pass. In MoE, this is demonstrated to improve performance with the same computational budget \cite{fedus2022switchtransformersscalingtrillion}. We compare different inference strategies empirically to outline this efficiency-correctness tradeoff, where we find selecting only the top expert is the most efficient approach and does not sacrifice quality.

\subsection{Distillation}

In production systems, it is difficult to deploy models with extremely high parameter counts. A DDM model with single-expert selection at test-time achieves nearly the same FLOPs per forward pass as a monolithic model, even though it has $|K|$ times more parameters. Still, it requires $|K|$ times more memory than the monolithic model to load in the first place. As model sizes scale past the memory budget of today's accelerators, this becomes a challenging distributed systems problem. While this is achievable in many cases, we present a distillation approach as a convenient alternative.

Specifically, the sparse DDM model can be distilled into a dense model. After the distillation, inference can be done just like any other diffusion model trained in a non-decentralized manner. Note that the goal of distillation here is to reduce $K$ experts into a single model, not to reduce the number of denoising steps, as done in other diffusion distillation works ~\cite{salimans2022progressivedistillationfastsampling, song2023consistencymodels, yin2024onestepdiffusiondistributionmatching}.  We follow a teacher-student training procedure \cite{hinton2015distillingknowledgeneuralnetwork}. Specifically, we replace the conditional flow $u_t(x_t|x_0)$ supervision target with a prediction from the teacher model:
$$\mathcal{L}_{\text{distill}}(\theta) = \mathbb{E}_{t,q(x_0),p_t(x_t|x_0)} \left\lVert \mathbf{v}_{\theta,t}(x_t) - \mathbf{v}_{\text{teacher},t}(x_t) \right\rVert^2.$$

\noindent In our case, we use the cluster label assigned to each data point to select a single teacher expert per training example.

\section{Experiments}
\label{sec:experiments}

We evaluate the effectiveness of 
decentralized diffusion models on the academic ImageNet dataset and a subset of the LAION \cite{schuhmann2022laion5bopenlargescaledataset} dataset filtered for aesthetic scores of 5 or higher, which more closely resembles real-world training scenarios.
Our goal is to analyze the design space of decentralized diffusion models applied to widely-adopted settings so that practitioners can confidently adapt the method to their needs. For this reason, we use standard architectures, hyperparameters and data when possible.

\subsection{Implementation Details}

\noindent\textbf{Dataset Partitioning} We partition the dataset in an efficient manner that facilitates learning the decentralized flow matching objective. %
Naive k-means scales quadratically with input data size, which is a non-starter for massive Internet datasets. Ma et al. \cite{ma2024modeclipdataexperts} propose a multi-stage algorithm that efficiently consolidates a large number of fine-grained clusters into a small number of partitions. We adopt this approach to focus our efforts on generative modeling.

Following~\cite{ma2024modeclipdataexperts}, we compute image features for the images in the dataset by pooling the output from DINOv2 \cite{oquab2024dinov2learningrobustvisual} to incorporate both pixel features and semantics, cluster these features to 1024 fine-grained centroids, and then further consolidate to $k$ coarse centroids. We assign each data point to the nearest of the coarse centroids to produce the final set of partitions. This partitioning process is computationally negligible compared to training.

\vspace{0.5em}\noindent\textbf{Evaluations}
We control for known sensitivities in FID calculations that arise from different implementations and evaluation splits. We designate a fixed evaluation set of 50,000 samples from each training dataset for computing features and evaluating all models. Our FID figures align well with published results and our consistent implementation provides precise relative comparisons between different approaches. This standardization eliminates confounding variables that often complicate comparisons of generative models.

\begin{figure*}[ht!]
    \centering

    \begin{subfigure}[t]{0.31\linewidth}
        \centering
        \includegraphics[width=\linewidth]{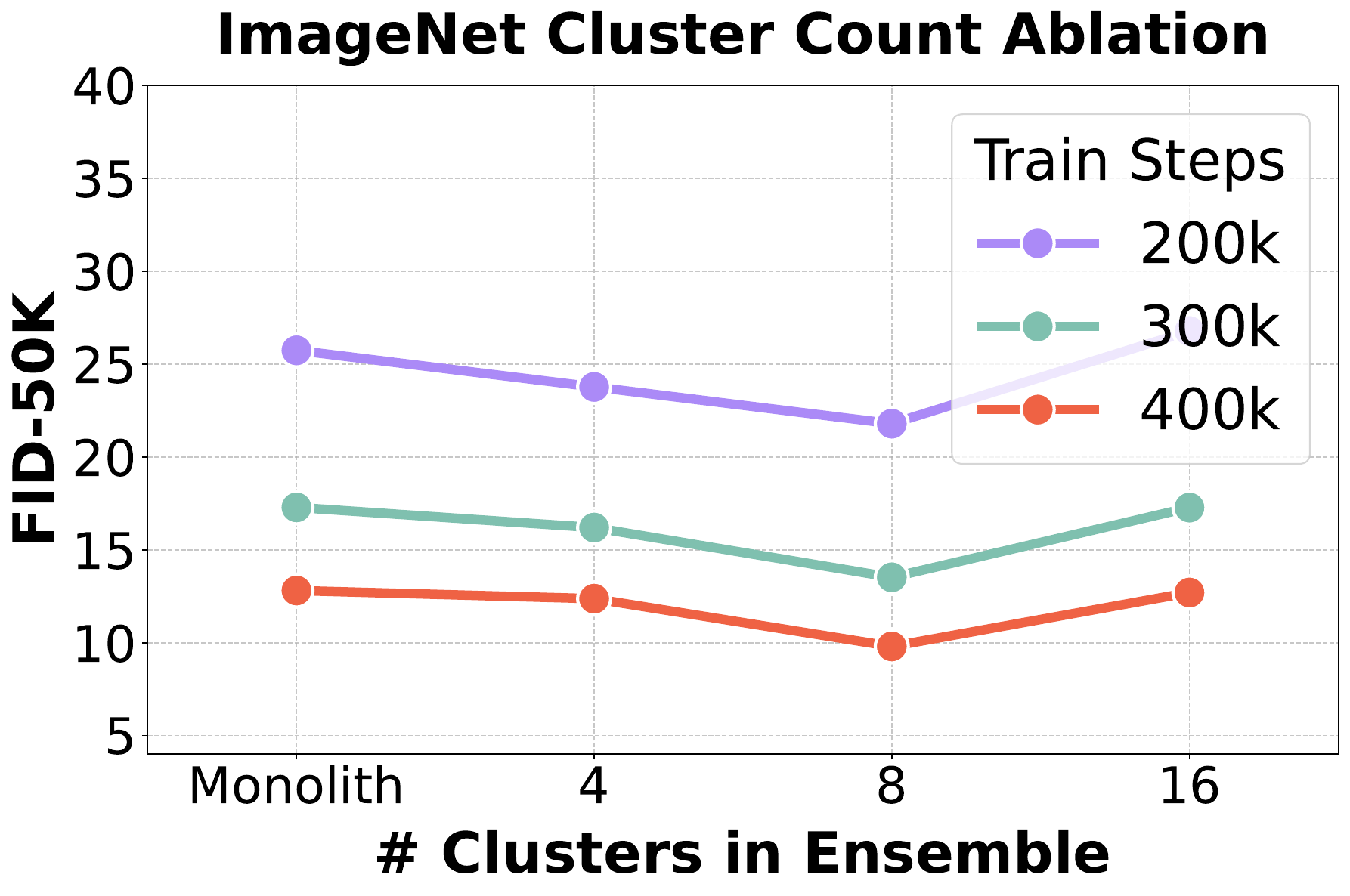}
          \vspace{-1.5em}
        \caption{}
        \label{fig:imagenet_num_clusters}
    \end{subfigure}
    \hfill
    \begin{subfigure}[t]{0.31\linewidth}
        \centering
        \includegraphics[width=\linewidth]{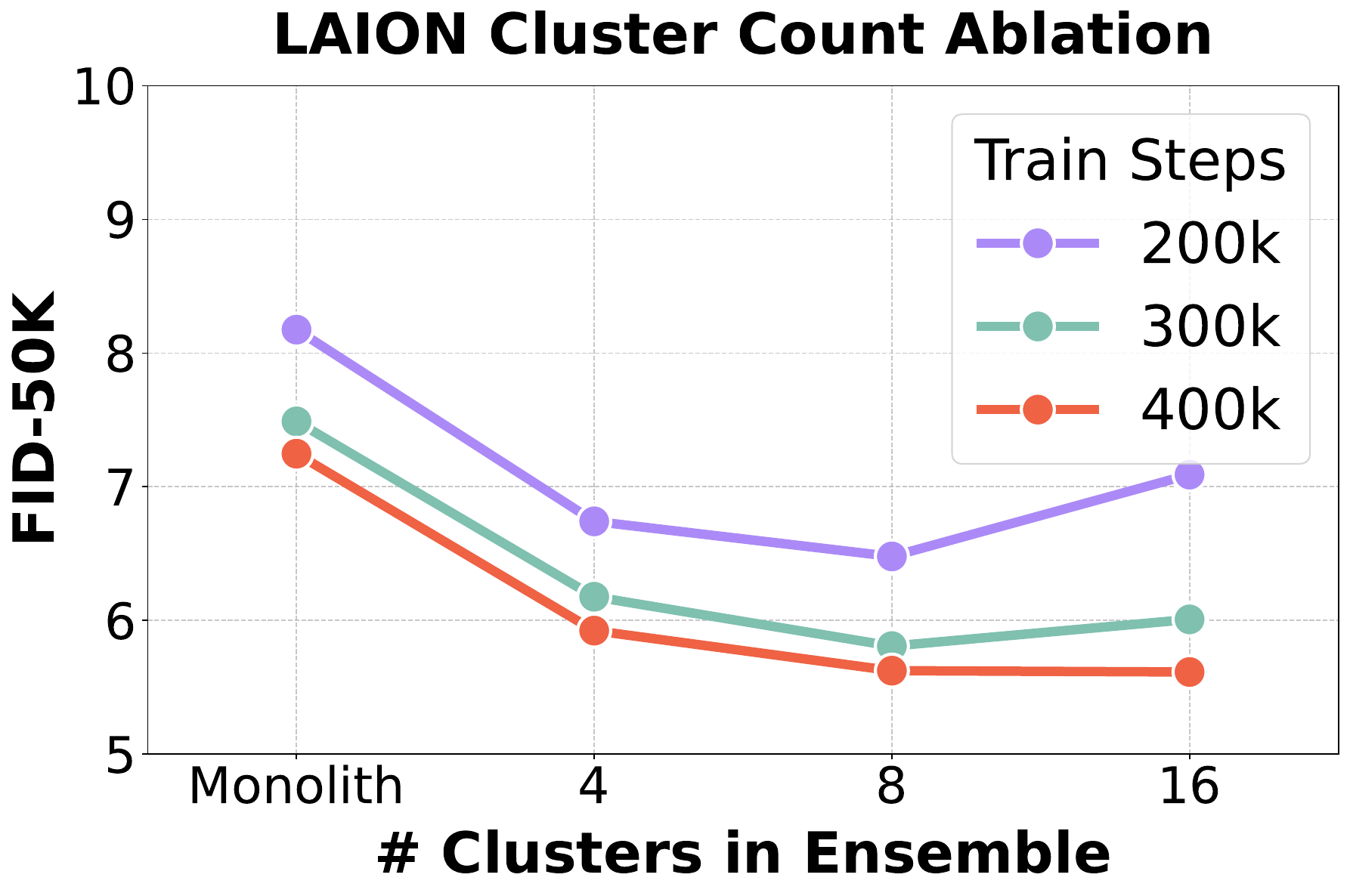}
          \vspace{-1.5em}
        \caption{}
        \label{fig:laion_num_clusters}
    \end{subfigure}
    \hfill
    \begin{subfigure}[t]{0.31\linewidth}
        \centering
        \includegraphics[width=\linewidth]{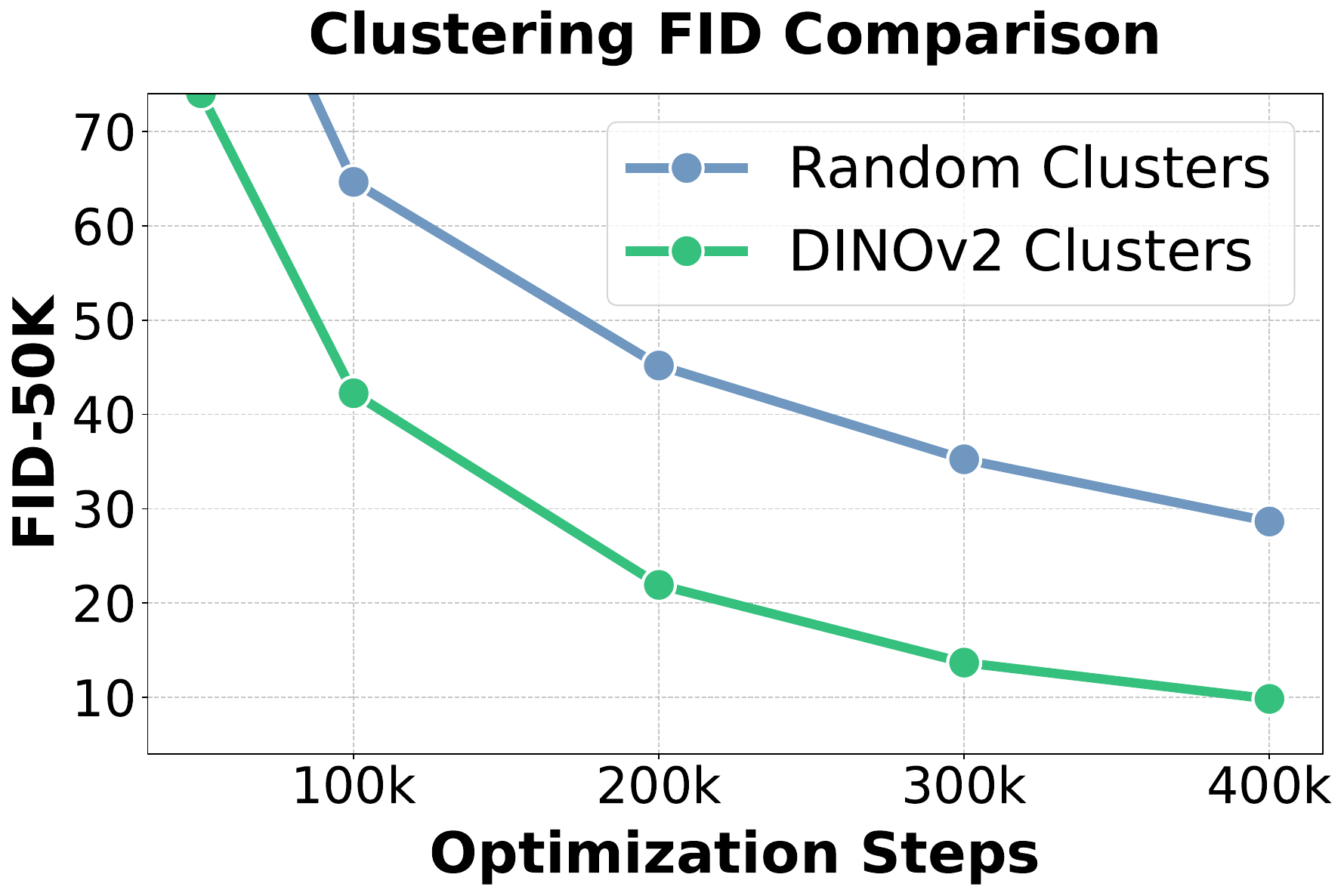}
          \vspace{-1.5em}
        \caption{}
        \label{fig:random_clusters}
    \end{subfigure}

    \begin{subfigure}[t]{0.38\linewidth}
        \centering
        \includegraphics[width=\linewidth]{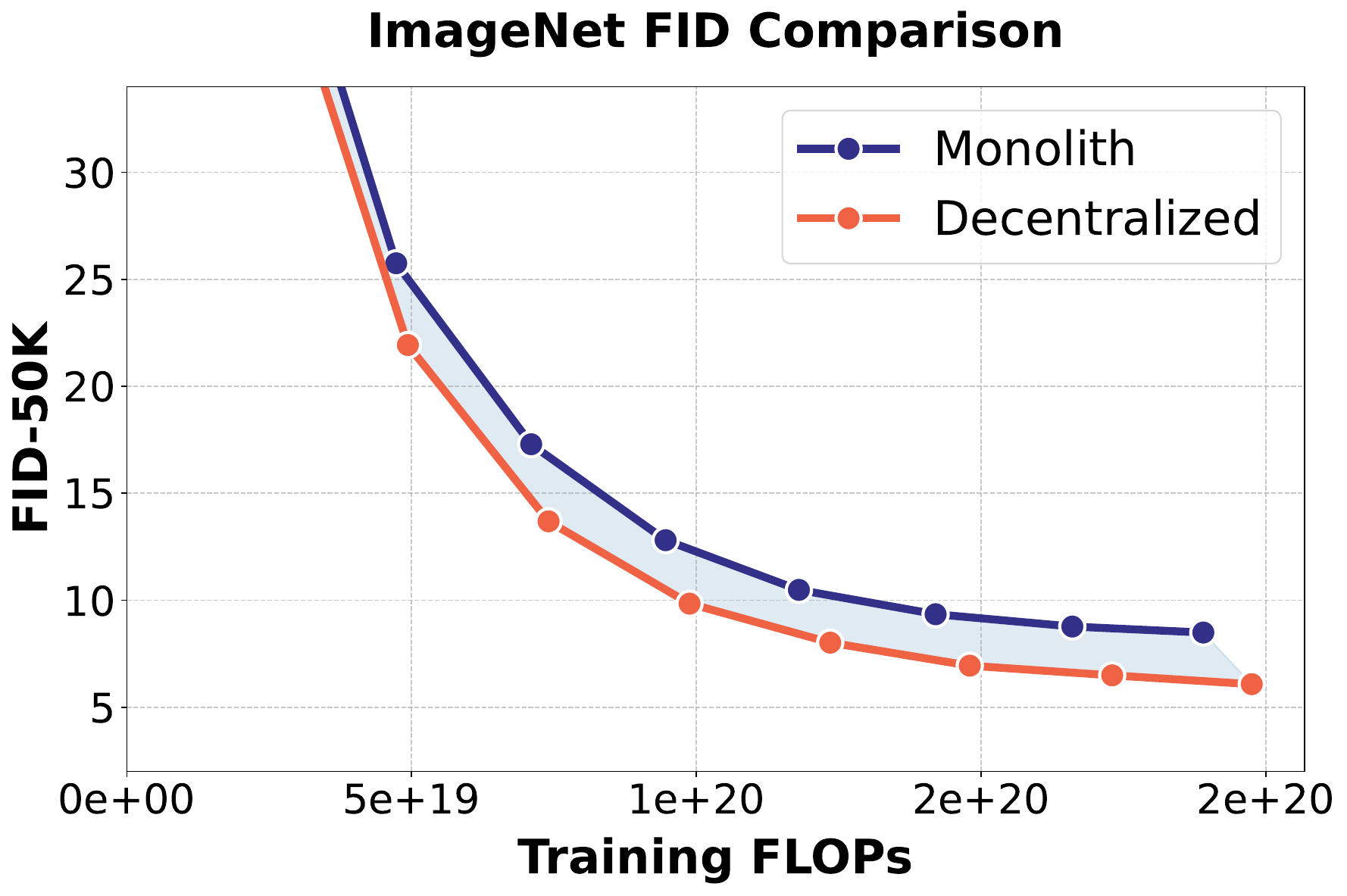}
          \vspace{-1.5em}
        \caption{}
        \label{fig:imagenet_beats_baseline}
    \end{subfigure}
    \hspace{10pt}
    \begin{subfigure}[t]{0.38\linewidth}
        \centering
        \includegraphics[width=\linewidth]{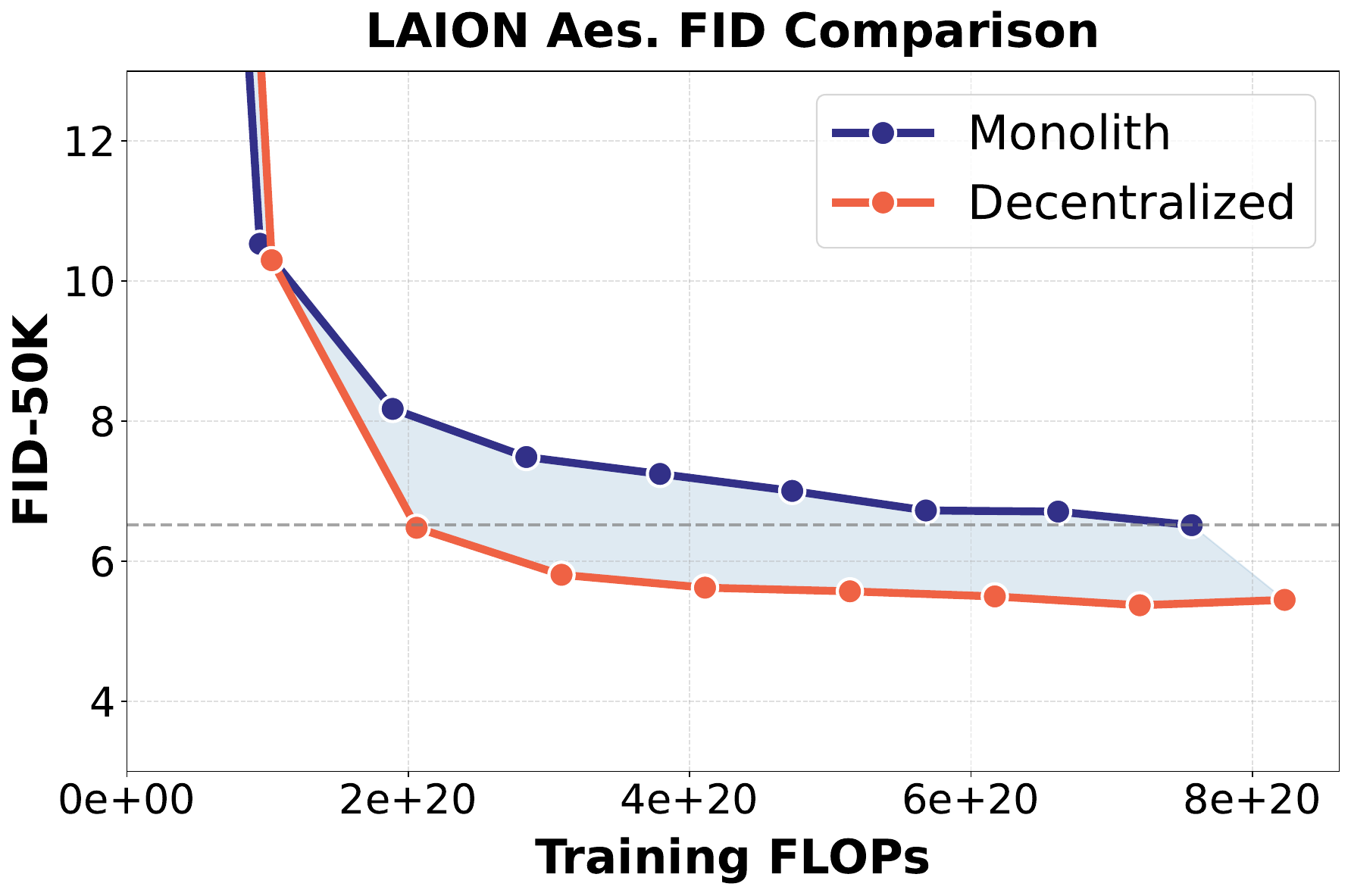}
          \vspace{-1.5em}
        \caption{}
        \label{fig:laion_beats_baseline}
    \end{subfigure}
    \vspace{-8pt}
    \caption{\textbf{Ablations at the DiT XL model scale.} Eight-expert DDMs display the best consistent performance on ImageNet (a) and LAION Aesthetics (b). We show the importance of image-based clustering on ImageNet compared to random clustering (c). Finally, FLOP-for-FLOP, decentralized diffusion models outperform monolith diffusion models on both datasets (d, e).}
    \label{fig:quantitative_comparisons}
    \vspace{-0.5em}
\end{figure*}

\begin{table}[!t]
\begin{center}

  \resizebox{0.9\columnwidth}{!}{
  \small
\begin{tabular}{@{}lcccc@{}}
\toprule
{\footnotesize Inference Strategy} & {\footnotesize GFLOPs $\downarrow$} & {\footnotesize FID $\downarrow$} & {\footnotesize CLIP FID $\downarrow$} \\ \midrule
Monolith & \textbf{308} & 12.81 & 5.58 \\
Oracle & \textbf{308} & 10.46 & 5.83 \\
Full & 2490 & 10.52 & 5.83  \\
Top-1 & 334 & \textbf{9.84} & \textbf{5.48}  \\
Top-2 & 642 & 10.31 & 5.74 \\
Top-3 & 950 & 10.37 & 5.77 \\
Sample-1 & 334 & 157.05 & 51.17  \\
Sample-2 & 642 & 10.27 & 5.73 \\
Sample-3 & 950 & 10.44 & 5.78 \\ 
Threshold-0.01 & - & 10.46 & 5.81 \\
Threshold-0.05 & - & 10.37 & 5.75 \\
Threshold-0.1 & - & 10.15 & 5.72 \\
Nucleus ($T=0.5$) & 334 & 188.66 & 60.09 \\
Nucleus ($T=1.0$) &  334 & 152.16 & 48.37 \\
Nucleus ($T=2.0$) & 334 & 33.9  & 14 \\
 \bottomrule
\end{tabular}
}

\vspace{-0.5em}
    \caption{\textbf{Test-Time Combination Strategies.} We ablate strategies to sample from the ensemble at test-time and find that simply selecting the top expert outperforms more sophisticated alternatives.}
    \vspace{-2em}
  \label{tab:emsemble}%
      
\end{center}
\end{table}%

\vspace{0.5em}\noindent\textbf{Training Details} For ImageNet experiments, we adopt the hyperparameters and architecture from diffusion transformers~\cite{peebles2023scalablediffusionmodelstransformers}, using a batch size of 256, EMA decay rate of 0.9999, and learning rate of 1e-4 without warmup or decay. We aim to replicate plausible real-world training runs with our LAION experiments, so we scale the batch size to 1024.

We reimplement the DiT XL/2 architecture for our denoising models, with each containing 895M parameters. In decentralized diffusion models, the total parameter count scales linearly with the number of experts. However, the computational cost remains constant during inference when using single expert selection.

For LAION experiments, we implement text conditioning following the Pix-Art Alpha~\cite{chen2023pixartalphafasttrainingdiffusion} architecture, using SDXL's CLIP~\cite{radford2021learningtransferablevisualmodels,Cherti_2023} model to incorporate text via cross-attention. The router uses the smaller DiT B/2 architecture (158M parameters) augmented with a learned CLS token that decodes linearly to a probability distribution over the clusters.

We ensure fair comparison by maintaining consistent total computation across decentralized diffusion models and baseline monolith models. We achieve this simply by dividing total batch size evenly among experts. For example, with eight experts, a 256 monolith batch size corresponds to eight expert batches of size 32. This equalizes the total training FLOPs between DDMs and baselines. The router introduces an additional 4\% measured training FLOPs overhead, which we also incorporate in our comparisons.

\subsection{Ensembling at Test-Time}

We first compare different strategies to combine expert predictions at test-time. A full estimate of the marginal flow involves linearly combining all expert predictions:

\begin{align}
    u_t(x_t)=\sum_{k=1}^K \underbrace{r_\theta(x_t,t)}_{\text{Router}} \underbrace{v_{\theta,t}(x_t)}_{\text{Expert}}.
\end{align}

\noindent In practice, selecting only important experts saves on computation. We evaluate the following strategies: 

\begin{itemize}
    \item \textbf{Full.} Compute the weighted combination of all expert predictions. This strategy's FLOP cost scales linearly with the number of experts.
    \item \textbf{Sample.} Sample from the router's predicted softmax distribution to select a single expert. This is an unbiased Monte-Carlo estimate of the marginal flow.
    \item \textbf{Top-k.} Simply use the $k$ experts with the maximum predicted router probability. Top-1 selection is the most efficient option at test-time.
    \item \textbf{Nucleus.} Sample one expert according to the nucleus (top-$p$) sampling strategy~\cite{holtzman2020curiouscaseneuraltext} commonly used in large language models. We use $p=0.9$ and ablate softmax temperature in Table~\ref{tab:emsemble}.
    \item \textbf{Oracle.} Select one expert according to the cluster label associated with an evaluation image. Used only to evaluate the effectiveness of the learned router.
\end{itemize}

\begin{figure}[t]
\centering
\includegraphics[width=\linewidth]{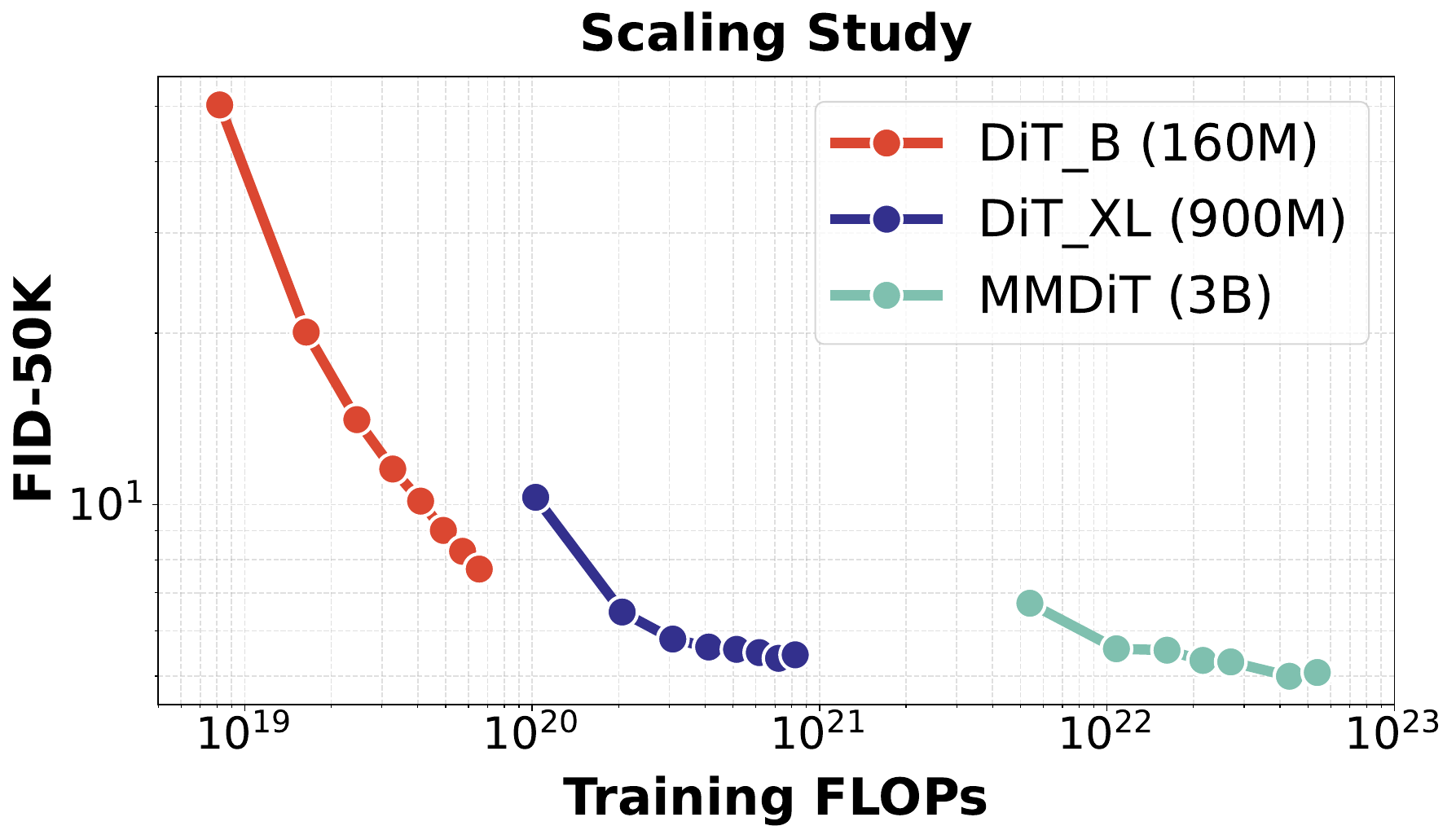}
        \caption{\textbf{Decentralized diffusion models scale gracefully to billions of parameters.} Throughout training, we plot the FID over LAION Aesthetics as a function of training compute. We find that increasing expert model capacity and training compute predictably improves performance.}
        \label{fig:laion_scaling}
\vspace{-0.5em}
\end{figure}

\noindent In Table~\ref{tab:emsemble}, we evaluate these inference strategies on ImageNet. We find that top-1 selection outperforms all other alternatives, while also incurring the lowest FLOP cost. For all other comparisons in Figures~\ref{fig:quantitative_comparisons} and~\ref{fig:laion_scaling}, we use top-1 selection because it nearly matches the computational cost of a dense model.

\subsection{Selecting the Right Number of Experts}

\begin{figure}[t]
\centering
\includegraphics[width=\linewidth]{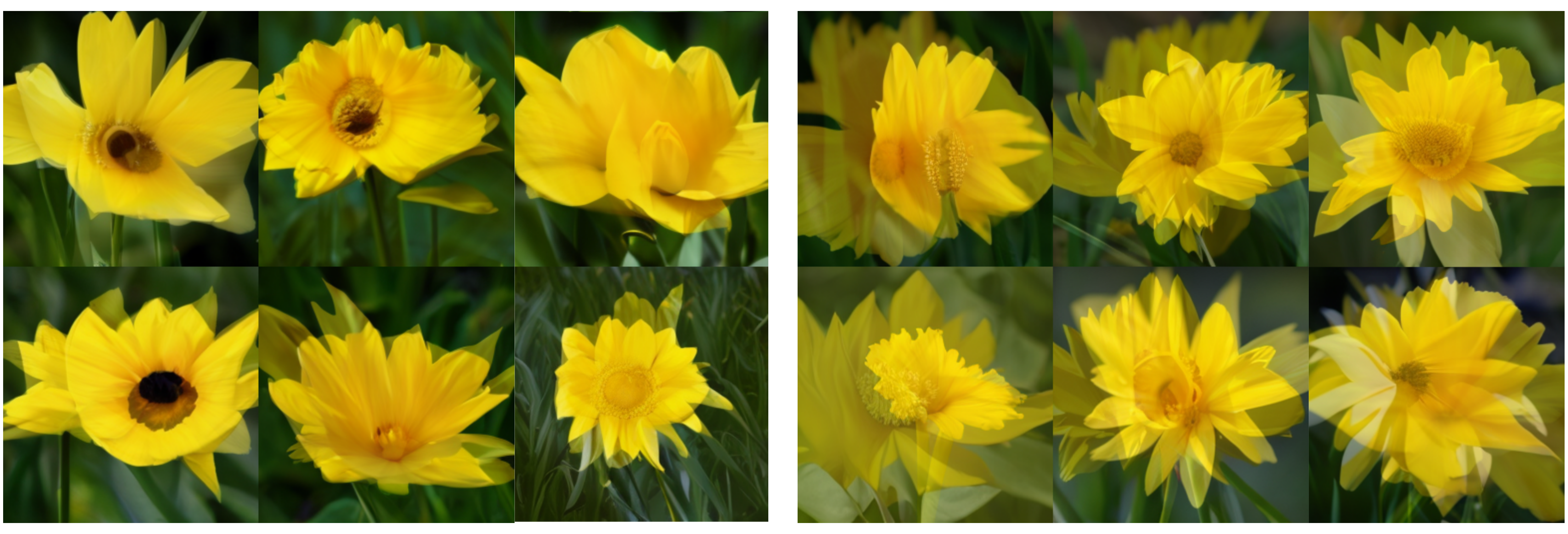}
\caption{\textbf{DDMs optimize the global diffusion objective.} We average samples from the monolithic and DDM ImageNet models using a deterministic sampler with matching random seeds (left) and compare them to outputs generated with random noise samples (right). The left samples are highly correlated, appearing less blurry.}
\label{fig:seed_match}
\vspace{-0.5em}
\end{figure}

\begin{figure}[t]
\centering
\includegraphics[width=\linewidth]{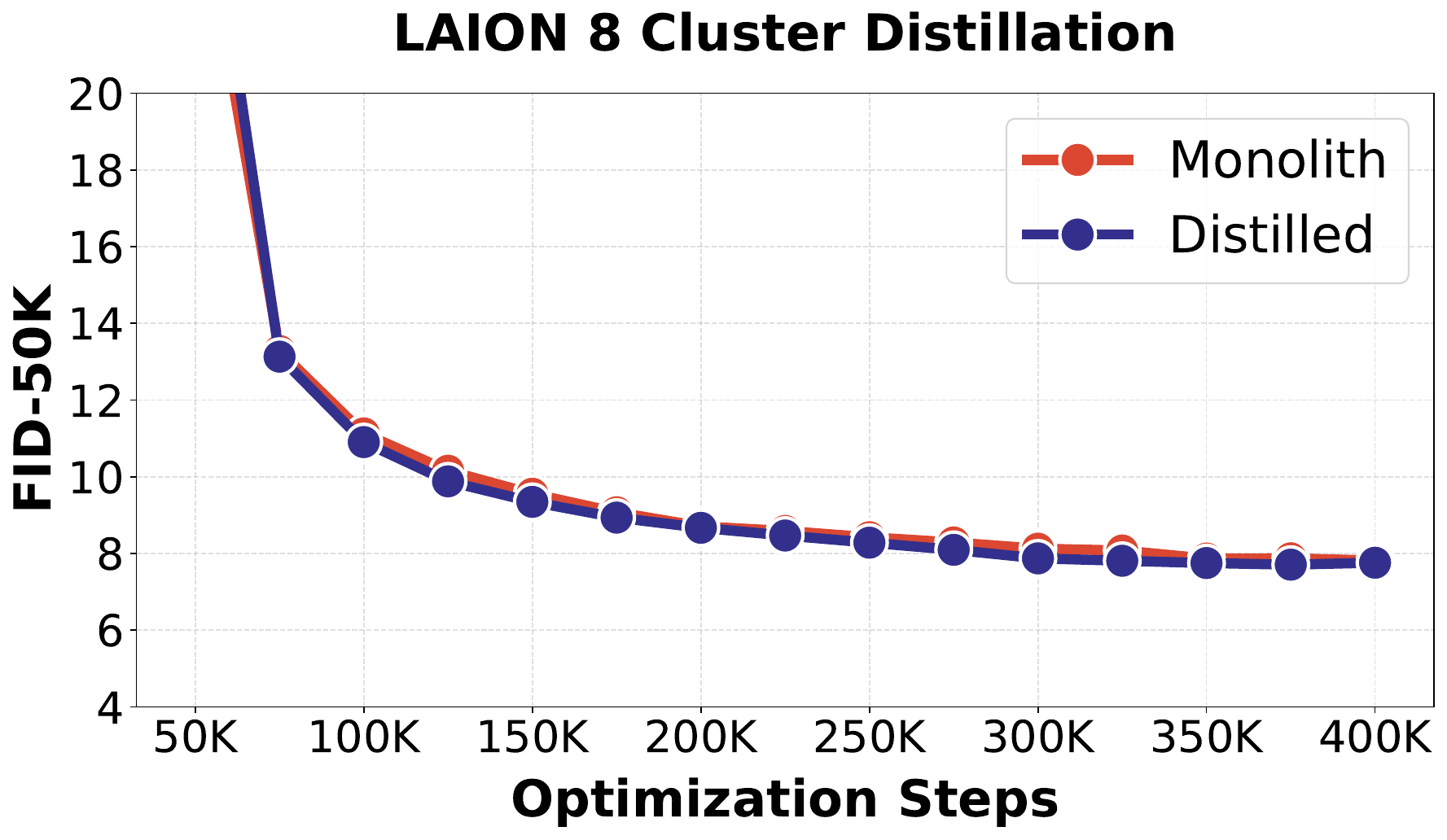}
\caption{\textbf{Distilling a DDM into a dense model.} Dense models are often more convenient than sparse models to serve in production settings. Distilling a decentralized diffusion model into a dense model matches the performance of training a monolith from scratch at one third of the FLOP-cost (1/4 batch size).}
\label{fig:distillation_main}
\vspace{-0.5em}
\end{figure}

The \method{} objective theoretically supports any number of experts, but practically we find this is an important hyperparameter for DDMs. This choice determines both the degree of decentralization and the total parameter count of the system. In the theoretical limit, where the number of experts approaches the number of training samples, the system would reduce to a nearest neighbor lookup that can only reproduce training examples—equivalent to the analytical form of flow matching. We also find that individual experts train poorly as the global batch size is divided too aggressively, as in the 16 expert (batch size 16 per expert) experiment in Figure~\ref{fig:imagenet_num_clusters}.

We compare DDMs with 4, 8 and 16 experts in Figures ~\ref{fig:imagenet_num_clusters} and~\ref{fig:laion_num_clusters}, and we find that eight experts achieves the best performance consistently on ImageNet and LAION. This configuration enables strong decentralization while maintaining reasonable test-time memory requirements. The eight experts appear to specialize meaningfully while preserving coherent coverage of the data distribution. Empirically, we find this is a sweet spot for the competing factors of model capacity, decentralization, and practical deployment.

\subsection{DDMs vs. Monoliths}
We compare decentralized diffusion models against a fair monolithic baseline on both datasets. We find that decentralized diffusion models with eight experts consistently outperform standard diffusion models on a FLOP-for-FLOP basis. In Figure~\ref{fig:imagenet_beats_baseline}, we compare FID on ImageNet up to 800k training steps. The decentralized diffusion model achieves a 28\% lower FID at 800k steps (6.081 vs. 8.494). Note that we plot FID as a function of training FLOPs to account for the 4\% additional training cost of the router.

We find that the performance uplift is also significant on captioned internet data with LAION in Figure~\ref{fig:laion_beats_baseline}. In fact, it achieves a lower FID of 6.48 at 200k optimization steps than the monolith's 6.52 FID at 800k steps. This represents a 4x training speedup as well as a lower convergence FID.

We also visually verify the correctness of the \method{} objective. The standard diffusion objective predictably pairs noise samples and data samples, so a well-trained DDM should sample a similar image as a monolith for the same input noise. We verify this in Figure~\ref{fig:seed_match}. Please see more samples and analysis of DDM in the supplemental.

\subsection{Data Clustering Ablation}

\method{} imposes no explicit constraints on cluster size or composition, so we ablate two clustering strategies. We find that the chosen strategy significantly impacts model performance. Comparing feature-based clustering using DINO against random cluster assignments, which would maintain i.i.d. properties across partitions, reveals that feature-based clustering improves results significantly. We hypothesize that there is more mutual information within feature clusters than random clusters, meaning expert models can more efficiently compress and specialize in their assigned sub-distributions. When data possesses semantic or low-level feature similarities, experts are free to learn more focused representations.

\subsection{Distillation}

While our method achieves computational efficiency through top-1 expert selection at inference time, the total memory footprint of many expert models can be substantial. We address this limitation through knowledge distillation, compressing the ensemble's capabilities into a dense model. Our approach supervises a student model with predictions from the expert ensemble, selecting the appropriate expert for each training example based on its cluster label. This can be seen as distilling the top-1 sparse model and incurs the same cost as standard teacher-student distillation.

Our distilled model matches the performance of directly training on the dataset, despite using only one quarter of the batch size and, consequently, one third the training FLOPs (assuming a backward pass costs double a forward pass). After 400k training steps, the distilled model achieves an FID of 7.76, comparable to the baseline model's FID of 7.82 (Figure~\ref{fig:distillation_main}). Many diffusion distillation works focus on reducing the number of sampling steps, whereas we just aim to replicate the ensemble in a dense model. We leave the exploration of combining our method with sampling-focused distillation techniques as promising future work.

\subsection{Scaling Experiment}

We perform a scaling study of decentralized diffusion models. At each scale, we follow best practices gleaned from our ablations and train a system of eight expert models, each based on the FLUX MMDiT architecture~\cite{esser2024scalingrectifiedflowtransformers}. We use a hidden dimension of 2560, depth of 30, and separate text and visual token streams, which total 3B parameters per expert. We encode text prompts using a single T5 XL model~\cite{raffel2023exploringlimitstransferlearning} and mix their features with image features through self attention.

Crucially, each expert can be trained independently on readily available hardware. With 16 GPUs per expert, we train at 0.28 seconds per iteration for 1M pretraining steps. Using gradient accumulation, this is equivalent to training each expert on a single on-demand cloud GPU node for six and a half days. This demonstrates that our method enables training large-scale diffusion models without specialized infrastructure or large integrated compute clusters.

We evaluate our large-scale ensemble against smaller DiT B and XL~\cite{peebles2023scalablediffusionmodelstransformers} ensembles in Figure~\ref{fig:laion_scaling}. DDM performance improves as a function of expert parametrization and does not saturate at any scale we tried. We finally finetune our largest ensemble for 60k steps on high-resolution data and display some selected samples in Figure~\ref{fig:full-page}.

\section{Discussion}

Decentralized diffusion models enable high-quality generative model training across isolated compute clusters, greatly broadening the possible hardware configurations for diffusion model training. While we focus on distributing computational resources, DFM theoretically permits decentralizing data as well—a property with potential privacy implications for domains like medical imaging.
Experts can train locally where sensitive data resides, and a router can train on samples from these experts rather than the raw data. These ideas allow DDM to preserve data privacy and sovereignty, as private data never leaves its original location. Furthermore, combining DDMs with low-bandwidth training methods could push the boundaries of decentralization—perhaps enabling large-scale model training on true commodity hardware. While we experiment on image modeling, the principles proposed by DDM may be applied to other domains, such as medical imaging, robotic policies and video modeling. We look forward to future works in these directions.

\paragraph{Acknowledgements.} We would like to thank Alex Yu for his guidance throughout the project and his score matching derivation. We would also like to thank Daniel Mendelevitch, Songwei Ge, Dan Kondratyuk, Haiwen Feng, Terrance DeVries, Chung Min Kim, Hang Gao, Justin Kerr and the Luma AI research team for helpful discussions.

{
    \small
    \bibliographystyle{ieeenat_fullname}
    \bibliography{main}

\begin{thebibliography}{54}
\providecommand{\natexlab}[1]{#1}
\providecommand{\url}[1]{\texttt{#1}}
\expandafter\ifx\csname urlstyle\endcsname\relax
  \providecommand{\doi}[1]{doi: #1}\else
  \providecommand{\doi}{doi: \begingroup \urlstyle{rm}\Url}\fi

\bibitem[An et~al.(2024)An, Bi, Chen, Chen, Deng, Ding, Dong, Du, Gao, Guan, et~al.]{an2024fireflyeraihpccosteffectivesoftwarehardware}
Wei An, Xiao Bi, Guanting Chen, Shanhuang Chen, Chengqi Deng, Honghui Ding, Kai Dong, Qiushi Du, Wenjun Gao, Kang Guan, et~al.
\newblock Fire-flyer ai-hpc: A cost-effective software-hardware co-design for deep learning.
\newblock \emph{arXiv preprint arXiv:2408.14158}, 2024.

\bibitem[Biggs et~al.(2024)Biggs, Seshadri, Zou, Jain, Golatkar, Xie, Achille, Swaminathan, and Soatto]{biggs2024diffusionsoupmodelmerging}
Benjamin Biggs, Arjun Seshadri, Yang Zou, Achin Jain, Aditya Golatkar, Yusheng Xie, Alessandro Achille, Ashwin Swaminathan, and Stefano Soatto.
\newblock Diffusion soup: Model merging for text-to-image diffusion models.
\newblock \emph{arXiv preprint arXiv:2406.08431}, 2024.

\bibitem[Bommasani et~al.(2021)Bommasani, Hudson, Adeli, Altman, Arora, von Arx, Bernstein, Bohg, Bosselut, Brunskill, et~al.]{bommasani2022opportunitiesrisksfoundationmodels}
Rishi Bommasani, Drew~A Hudson, Ehsan Adeli, Russ Altman, Simran Arora, Sydney von Arx, Michael~S Bernstein, Jeannette Bohg, Antoine Bosselut, Emma Brunskill, et~al.
\newblock On the opportunities and risks of foundation models.
\newblock \emph{arXiv preprint arXiv:2108.07258}, 2021.

\bibitem[Brooks et~al.(2024)Brooks, Peebles, Holmes, DePue, Guo, Jing, Schnurr, Taylor, Luhman, Luhman, et~al.]{videoworldsimulators2024}
Tim Brooks, Bill Peebles, Connor Holmes, Will DePue, Yufei Guo, Li Jing, David Schnurr, Joe Taylor, Troy Luhman, Eric Luhman, et~al.
\newblock Video generation models as world simulators. 2024.
\newblock \emph{URL https://openai. com/research/video-generation-models-as-world-simulators}, 3, 2024.

\bibitem[Brown et~al.(2022)Brown, Caterini, Ross, Cresswell, and Loaiza-Ganem]{brown2023verifyingunionmanifoldshypothesis}
Bradley~CA Brown, Anthony~L Caterini, Brendan~Leigh Ross, Jesse~C Cresswell, and Gabriel Loaiza-Ganem.
\newblock Verifying the union of manifolds hypothesis for image data.
\newblock \emph{arXiv preprint arXiv:2207.02862}, 2022.

\bibitem[Brown et~al.(2020)Brown, Mann, Ryder, Subbiah, Kaplan, Dhariwal, Neelakantan, Shyam, Sastry, Askell, and et~al.]{brown2020languagemodelsfewshotlearners}
Tom~B. Brown, Benjamin Mann, Nick Ryder, Melanie Subbiah, Jared Kaplan, Prafulla Dhariwal, Arvind Neelakantan, Pranav Shyam, Girish Sastry, Amanda Askell, and Sandhini~Agarwal et al.
\newblock Language models are few-shot learners, 2020.

\bibitem[Chen et~al.(2023)Chen, Yu, Ge, Yao, Xie, Wu, Wang, Kwok, Luo, Lu, and Li]{chen2023pixartalphafasttrainingdiffusion}
Junsong Chen, Jincheng Yu, Chongjian Ge, Lewei Yao, Enze Xie, Yue Wu, Zhongdao Wang, James Kwok, Ping Luo, Huchuan Lu, and Zhenguo Li.
\newblock Pixart-$\alpha$: Fast training of diffusion transformer for photorealistic text-to-image synthesis, 2023.

\bibitem[Chen et~al.(2018)Chen, Rubanova, Bettencourt, and Duvenaud]{chenNODE}
Ricky~TQ Chen, Yulia Rubanova, Jesse Bettencourt, and David~K Duvenaud.
\newblock Neural ordinary differential equations.
\newblock \emph{Advances in neural information processing systems}, 31, 2018.

\bibitem[Cherti et~al.(2023)Cherti, Beaumont, Wightman, Wortsman, Ilharco, Gordon, Schuhmann, Schmidt, and Jitsev]{Cherti_2023}
Mehdi Cherti, Romain Beaumont, Ross Wightman, Mitchell Wortsman, Gabriel Ilharco, Cade Gordon, Christoph Schuhmann, Ludwig Schmidt, and Jenia Jitsev.
\newblock Reproducible scaling laws for contrastive language-image learning.
\newblock In \emph{Proceedings of the IEEE/CVF Conference on Computer Vision and Pattern Recognition}, pages 2818--2829, 2023.

\bibitem[Chi et~al.(2023)Chi, Xu, Feng, Cousineau, Du, Burchfiel, Tedrake, and Song]{chi2024diffusionpolicyvisuomotorpolicy}
Cheng Chi, Zhenjia Xu, Siyuan Feng, Eric Cousineau, Yilun Du, Benjamin Burchfiel, Russ Tedrake, and Shuran Song.
\newblock Diffusion policy: Visuomotor policy learning via action diffusion.
\newblock \emph{The International Journal of Robotics Research}, page 02783649241273668, 2023.

\bibitem[DeepSeek-AI et~al.(2024)DeepSeek-AI, Liu, Feng, Xue, Wang, Wu, Lu, Zhao, Deng, Zhang, Ruan, Dai, Guo, Yang, Chen, Ji, Li, Lin, Dai, Luo, Hao, Chen, Li, Zhang, Bao, Xu, Wang, Zhang, Ding, Xin, Gao, Li, Qu, Cai, Liang, Guo, Ni, Li, Wang, Chen, Chen, Yuan, Qiu, Li, Song, Dong, Hu, Gao, Guan, Huang, Yu, Wang, Zhang, Xu, Xia, Zhao, Wang, Zhang, Li, Wang, Zhang, Zhang, Tang, Li, Tian, Huang, Wang, Zhang, Wang, Zhu, Chen, Du, Chen, Jin, Ge, Zhang, Pan, Wang, Xu, Zhang, Chen, Li, Lu, Zhou, Chen, Wu, Ye, Ye, Ma, Wang, Zhou, Yu, Zhou, Pan, Wang, Yun, Pei, Sun, Xiao, Zeng, Zhao, An, Liu, Liang, Gao, Yu, Zhang, Li, Jin, Wang, Bi, Liu, Wang, Shen, Chen, Zhang, Chen, Nie, Sun, Wang, Cheng, Liu, Xie, Liu, Yu, Song, Shan, Zhou, Yang, Li, Su, Lin, Li, Wang, Wei, Zhu, Zhang, Xu, Xu, Huang, Li, Zhao, Sun, Li, Wang, Yu, Zheng, Zhang, Shi, Xiong, He, Tang, Piao, Wang, Tan, Ma, Liu, Guo, Wu, Ou, Zhu, Wang, Gong, Zou, He, Zha, Xiong, Ma, Yan, Luo, You, Liu, Zhou, Wu, Ren, Ren, Sha, Fu, Xu, Huang, Zhang, Xie, Zhang, Hao,
  Gou, Ma, Yan, Shao, Xu, Wu, Zhang, Li, Gu, Zhu, Liu, Li, Xie, Song, Gao, and Pan]{deepseekai2024deepseekv3technicalreport}
DeepSeek-AI, Aixin Liu, Bei Feng, Bing Xue, Bingxuan Wang, Bochao Wu, Chengda Lu, Chenggang Zhao, Chengqi Deng, Chenyu Zhang, Chong Ruan, Damai Dai, Daya Guo, Dejian Yang, Deli Chen, Dongjie Ji, Erhang Li, Fangyun Lin, Fucong Dai, Fuli Luo, Guangbo Hao, Guanting Chen, Guowei Li, H. Zhang, Han Bao, Hanwei Xu, Haocheng Wang, Haowei Zhang, Honghui Ding, Huajian Xin, Huazuo Gao, Hui Li, Hui Qu, J.~L. Cai, Jian Liang, Jianzhong Guo, Jiaqi Ni, Jiashi Li, Jiawei Wang, Jin Chen, Jingchang Chen, Jingyang Yuan, Junjie Qiu, Junlong Li, Junxiao Song, Kai Dong, Kai Hu, Kaige Gao, Kang Guan, Kexin Huang, Kuai Yu, Lean Wang, Lecong Zhang, Lei Xu, Leyi Xia, Liang Zhao, Litong Wang, Liyue Zhang, Meng Li, Miaojun Wang, Mingchuan Zhang, Minghua Zhang, Minghui Tang, Mingming Li, Ning Tian, Panpan Huang, Peiyi Wang, Peng Zhang, Qiancheng Wang, Qihao Zhu, Qinyu Chen, Qiushi Du, R.~J. Chen, R.~L. Jin, Ruiqi Ge, Ruisong Zhang, Ruizhe Pan, Runji Wang, Runxin Xu, Ruoyu Zhang, Ruyi Chen, S.~S. Li, Shanghao Lu, Shangyan Zhou, Shanhuang
  Chen, Shaoqing Wu, Shengfeng Ye, Shengfeng Ye, Shirong Ma, Shiyu Wang, Shuang Zhou, Shuiping Yu, Shunfeng Zhou, Shuting Pan, T. Wang, Tao Yun, Tian Pei, Tianyu Sun, W.~L. Xiao, Wangding Zeng, Wanjia Zhao, Wei An, Wen Liu, Wenfeng Liang, Wenjun Gao, Wenqin Yu, Wentao Zhang, X.~Q. Li, Xiangyue Jin, Xianzu Wang, Xiao Bi, Xiaodong Liu, Xiaohan Wang, Xiaojin Shen, Xiaokang Chen, Xiaokang Zhang, Xiaosha Chen, Xiaotao Nie, Xiaowen Sun, Xiaoxiang Wang, Xin Cheng, Xin Liu, Xin Xie, Xingchao Liu, Xingkai Yu, Xinnan Song, Xinxia Shan, Xinyi Zhou, Xinyu Yang, Xinyuan Li, Xuecheng Su, Xuheng Lin, Y.~K. Li, Y.~Q. Wang, Y.~X. Wei, Y.~X. Zhu, Yang Zhang, Yanhong Xu, Yanhong Xu, Yanping Huang, Yao Li, Yao Zhao, Yaofeng Sun, Yaohui Li, Yaohui Wang, Yi Yu, Yi Zheng, Yichao Zhang, Yifan Shi, Yiliang Xiong, Ying He, Ying Tang, Yishi Piao, Yisong Wang, Yixuan Tan, Yiyang Ma, Yiyuan Liu, Yongqiang Guo, Yu Wu, Yuan Ou, Yuchen Zhu, Yuduan Wang, Yue Gong, Yuheng Zou, Yujia He, Yukun Zha, Yunfan Xiong, Yunxian Ma, Yuting Yan, Yuxiang
  Luo, Yuxiang You, Yuxuan Liu, Yuyang Zhou, Z.~F. Wu, Z.~Z. Ren, Zehui Ren, Zhangli Sha, Zhe Fu, Zhean Xu, Zhen Huang, Zhen Zhang, Zhenda Xie, Zhengyan Zhang, Zhewen Hao, Zhibin Gou, Zhicheng Ma, Zhigang Yan, Zhihong Shao, Zhipeng Xu, Zhiyu Wu, Zhongyu Zhang, Zhuoshu Li, Zihui Gu, Zijia Zhu, Zijun Liu, Zilin Li, Ziwei Xie, Ziyang Song, Ziyi Gao, and Zizheng Pan.
\newblock Deepseek-v3 technical report, 2024.

\bibitem[Devlin et~al.(2018)Devlin, Chang, Lee, and Toutanova]{devlin2019bertpretrainingdeepbidirectional}
Jacob Devlin, Ming-Wei Chang, Kenton Lee, and Kristina Toutanova.
\newblock Bert: Pre-training of deep bidirectional transformers for language understanding.
\newblock \emph{arXiv preprint arXiv:1810.04805}, 2018.

\bibitem[Douillard et~al.(2023)Douillard, Feng, Rusu, Chhaparia, Donchev, Kuncoro, Ranzato, Szlam, and Shen]{douillard2024dilocodistributedlowcommunicationtraining}
Arthur Douillard, Qixuan Feng, Andrei~A Rusu, Rachita Chhaparia, Yani Donchev, Adhiguna Kuncoro, Marc'Aurelio Ranzato, Arthur Szlam, and Jiajun Shen.
\newblock Diloco: Distributed low-communication training of language models.
\newblock \emph{arXiv preprint arXiv:2311.08105}, 2023.

\bibitem[Dubey et~al.(2024)Dubey, Jauhri, Pandey, Kadian, Al-Dahle, Letman, Mathur, Schelten, Yang, Fan, et~al.]{dubey2024llama3herdmodels}
Abhimanyu Dubey, Abhinav Jauhri, Abhinav Pandey, Abhishek Kadian, Ahmad Al-Dahle, Aiesha Letman, Akhil Mathur, Alan Schelten, Amy Yang, Angela Fan, et~al.
\newblock The llama 3 herd of models.
\newblock \emph{arXiv preprint arXiv:2407.21783}, 2024.

\bibitem[Esser et~al.(2024)Esser, Kulal, Blattmann, Entezari, M{\"u}ller, Saini, Levi, Lorenz, Sauer, Boesel, et~al.]{esser2024scalingrectifiedflowtransformers}
Patrick Esser, Sumith Kulal, Andreas Blattmann, Rahim Entezari, Jonas M{\"u}ller, Harry Saini, Yam Levi, Dominik Lorenz, Axel Sauer, Frederic Boesel, et~al.
\newblock Scaling rectified flow transformers for high-resolution image synthesis.
\newblock In \emph{Forty-first International Conference on Machine Learning}, 2024.

\bibitem[Fedus et~al.(2022)Fedus, Zoph, and Shazeer]{fedus2022switchtransformersscalingtrillion}
William Fedus, Barret Zoph, and Noam Shazeer.
\newblock Switch transformers: Scaling to trillion parameter models with simple and efficient sparsity.
\newblock \emph{Journal of Machine Learning Research}, 23\penalty0 (120):\penalty0 1--39, 2022.

\bibitem[Gale et~al.(2023)Gale, Narayanan, Young, and Zaharia]{gale2022megablocksefficientsparsetraining}
Trevor Gale, Deepak Narayanan, Cliff Young, and Matei Zaharia.
\newblock Megablocks: Efficient sparse training with mixture-of-experts.
\newblock \emph{Proceedings of Machine Learning and Systems}, 5:\penalty0 288--304, 2023.

\bibitem[Golatkar et~al.(2024)Golatkar, Achille, Swaminathan, and Soatto]{golatkar2024trainingdataprotectioncompositional}
Aditya Golatkar, Alessandro Achille, Ashwin Swaminathan, and Stefano Soatto.
\newblock Training data protection with compositional diffusion models, 2024.

\bibitem[Hinton(2015)]{hinton2015distillingknowledgeneuralnetwork}
Geoffrey Hinton.
\newblock Distilling the knowledge in a neural network.
\newblock \emph{arXiv preprint arXiv:1503.02531}, 2015.

\bibitem[Ho and Salimans(2022)]{ho2022classifierfreediffusionguidance}
Jonathan Ho and Tim Salimans.
\newblock Classifier-free diffusion guidance, 2022.

\bibitem[Ho et~al.(2020)Ho, Jain, and Abbeel]{ho2020denoising}
Jonathan Ho, Ajay Jain, and Pieter Abbeel.
\newblock Denoising diffusion probabilistic models.
\newblock \emph{Advances in neural information processing systems}, 33:\penalty0 6840--6851, 2020.

\bibitem[Holtzman et~al.(2020)Holtzman, Buys, Du, Forbes, and Choi]{holtzman2020curiouscaseneuraltext}
Ari Holtzman, Jan Buys, Li Du, Maxwell Forbes, and Yejin Choi.
\newblock The curious case of neural text degeneration, 2020.

\bibitem[Hoogeboom et~al.(2023)Hoogeboom, Heek, and Salimans]{hoogeboom2023simplediffusionendtoenddiffusion}
Emiel Hoogeboom, Jonathan Heek, and Tim Salimans.
\newblock Simple diffusion: End-to-end diffusion for high resolution images, 2023.

\bibitem[Jiang et~al.(2024)Jiang, Sablayrolles, Roux, Mensch, Savary, Bamford, Chaplot, Casas, Hanna, Bressand, et~al.]{jiang2024mixtralexperts}
Albert~Q Jiang, Alexandre Sablayrolles, Antoine Roux, Arthur Mensch, Blanche Savary, Chris Bamford, Devendra~Singh Chaplot, Diego de~las Casas, Emma~Bou Hanna, Florian Bressand, et~al.
\newblock Mixtral of experts.
\newblock \emph{arXiv preprint arXiv:2401.04088}, 2024.

\bibitem[Kamkari et~al.(2024)Kamkari, Ross, Hosseinzadeh, Cresswell, and Loaiza-Ganem]{kamkari2024geometricviewdatacomplexity}
Hamidreza Kamkari, Brendan~Leigh Ross, Rasa Hosseinzadeh, Jesse~C Cresswell, and Gabriel Loaiza-Ganem.
\newblock A geometric view of data complexity: Efficient local intrinsic dimension estimation with diffusion models.
\newblock \emph{arXiv preprint arXiv:2406.03537}, 2024.

\bibitem[Li et~al.(2022)Li, Gururangan, Dettmers, Lewis, Althoff, Smith, and Zettlemoyer]{li2022branchtrainmergeembarrassinglyparalleltraining}
Margaret Li, Suchin Gururangan, Tim Dettmers, Mike Lewis, Tim Althoff, Noah~A Smith, and Luke Zettlemoyer.
\newblock Branch-train-merge: Embarrassingly parallel training of expert language models.
\newblock \emph{arXiv preprint arXiv:2208.03306}, 2022.

\bibitem[Lipman et~al.(2022)Lipman, Chen, Ben-Hamu, Nickel, and Le]{lipman2023flowmatchinggenerativemodeling}
Yaron Lipman, Ricky~TQ Chen, Heli Ben-Hamu, Maximilian Nickel, and Matt Le.
\newblock Flow matching for generative modeling.
\newblock \emph{arXiv preprint arXiv:2210.02747}, 2022.

\bibitem[Liu et~al.(2022)Liu, Gong, and Liu]{liu2022flowstraightfastlearning}
Xingchao Liu, Chengyue Gong, and Qiang Liu.
\newblock Flow straight and fast: Learning to generate and transfer data with rectified flow.
\newblock \emph{arXiv preprint arXiv:2209.03003}, 2022.

\bibitem[Ma et~al.(2024)Ma, Huang, Xie, Li, Zettlemoyer, Chang, Yih, and Xu]{ma2024modeclipdataexperts}
Jiawei Ma, Po-Yao Huang, Saining Xie, Shang-Wen Li, Luke Zettlemoyer, Shih-Fu Chang, Wen-Tau Yih, and Hu Xu.
\newblock Mode: Clip data experts via clustering.
\newblock In \emph{Proceedings of the IEEE/CVF Conference on Computer Vision and Pattern Recognition}, pages 26354--26363, 2024.

\bibitem[McMahan et~al.(2017)McMahan, Moore, Ramage, Hampson, and y~Arcas]{mcmahan2023communicationefficientlearningdeepnetworks}
Brendan McMahan, Eider Moore, Daniel Ramage, Seth Hampson, and Blaise~Aguera y Arcas.
\newblock Communication-efficient learning of deep networks from decentralized data.
\newblock In \emph{Artificial intelligence and statistics}, pages 1273--1282. PMLR, 2017.

\bibitem[Morey(2024)]{Morey_2024}
Mark Morey.
\newblock Data center owners turn to nuclear as potential electricity source - u.s. energy information administration (eia), 2024.

\bibitem[Oquab et~al.(2023)Oquab, Darcet, Moutakanni, Vo, Szafraniec, Khalidov, Fernandez, Haziza, Massa, El-Nouby, et~al.]{oquab2024dinov2learningrobustvisual}
Maxime Oquab, Timoth{\'e}e Darcet, Th{\'e}o Moutakanni, Huy Vo, Marc Szafraniec, Vasil Khalidov, Pierre Fernandez, Daniel Haziza, Francisco Massa, Alaaeldin El-Nouby, et~al.
\newblock Dinov2: Learning robust visual features without supervision.
\newblock \emph{arXiv preprint arXiv:2304.07193}, 2023.

\bibitem[Peebles and Xie(2023)]{peebles2023scalablediffusionmodelstransformers}
William Peebles and Saining Xie.
\newblock Scalable diffusion models with transformers.
\newblock In \emph{Proceedings of the IEEE/CVF International Conference on Computer Vision}, pages 4195--4205, 2023.

\bibitem[Podell et~al.(2023)Podell, English, Lacey, Blattmann, Dockhorn, M{\"u}ller, Penna, and Rombach]{podell2023sdxlimprovinglatentdiffusion}
Dustin Podell, Zion English, Kyle Lacey, Andreas Blattmann, Tim Dockhorn, Jonas M{\"u}ller, Joe Penna, and Robin Rombach.
\newblock Sdxl: Improving latent diffusion models for high-resolution image synthesis.
\newblock \emph{arXiv preprint arXiv:2307.01952}, 2023.

\bibitem[Polyak et~al.(2024)Polyak, Zohar, Brown, Tjandra, Sinha, Lee, Vyas, Shi, Ma, Chuang, et~al.]{polyak2024moviegencastmedia}
Adam Polyak, Amit Zohar, Andrew Brown, Andros Tjandra, Animesh Sinha, Ann Lee, Apoorv Vyas, Bowen Shi, Chih-Yao Ma, Ching-Yao Chuang, et~al.
\newblock Movie gen: A cast of media foundation models.
\newblock \emph{arXiv preprint arXiv:2410.13720}, 2024.

\bibitem[Radford et~al.(2021)Radford, Kim, Hallacy, Ramesh, Goh, Agarwal, Sastry, Askell, Mishkin, Clark, et~al.]{radford2021learningtransferablevisualmodels}
Alec Radford, Jong~Wook Kim, Chris Hallacy, Aditya Ramesh, Gabriel Goh, Sandhini Agarwal, Girish Sastry, Amanda Askell, Pamela Mishkin, Jack Clark, et~al.
\newblock Learning transferable visual models from natural language supervision.
\newblock In \emph{International conference on machine learning}, pages 8748--8763. PMLR, 2021.

\bibitem[Raffel et~al.(2020)Raffel, Shazeer, Roberts, Lee, Narang, Matena, Zhou, Li, and Liu]{raffel2023exploringlimitstransferlearning}
Colin Raffel, Noam Shazeer, Adam Roberts, Katherine Lee, Sharan Narang, Michael Matena, Yanqi Zhou, Wei Li, and Peter~J Liu.
\newblock Exploring the limits of transfer learning with a unified text-to-text transformer.
\newblock \emph{Journal of machine learning research}, 21\penalty0 (140):\penalty0 1--67, 2020.

\bibitem[Reddi et~al.(2020)Reddi, Charles, Zaheer, Garrett, Rush, Kone{\v{c}}n{\`y}, Kumar, and McMahan]{reddi2021adaptivefederatedoptimization}
Sashank Reddi, Zachary Charles, Manzil Zaheer, Zachary Garrett, Keith Rush, Jakub Kone{\v{c}}n{\`y}, Sanjiv Kumar, and H~Brendan McMahan.
\newblock Adaptive federated optimization.
\newblock \emph{arXiv preprint arXiv:2003.00295}, 2020.

\bibitem[Rombach et~al.(2022)Rombach, Blattmann, Lorenz, Esser, and Ommer]{rombach2022highresolutionimagesynthesislatent}
Robin Rombach, Andreas Blattmann, Dominik Lorenz, Patrick Esser, and Bj{\"o}rn Ommer.
\newblock High-resolution image synthesis with latent diffusion models.
\newblock In \emph{Proceedings of the IEEE/CVF conference on computer vision and pattern recognition}, pages 10684--10695, 2022.

\bibitem[Saharia et~al.(2022)Saharia, Chan, Saxena, Li, Whang, Denton, Ghasemipour, Gontijo~Lopes, Karagol~Ayan, Salimans, et~al.]{saharia2022photorealistictexttoimagediffusionmodels}
Chitwan Saharia, William Chan, Saurabh Saxena, Lala Li, Jay Whang, Emily~L Denton, Kamyar Ghasemipour, Raphael Gontijo~Lopes, Burcu Karagol~Ayan, Tim Salimans, et~al.
\newblock Photorealistic text-to-image diffusion models with deep language understanding.
\newblock \emph{Advances in neural information processing systems}, 35:\penalty0 36479--36494, 2022.

\bibitem[Salimans and Ho(2022)]{salimans2022progressivedistillationfastsampling}
Tim Salimans and Jonathan Ho.
\newblock Progressive distillation for fast sampling of diffusion models.
\newblock \emph{arXiv preprint arXiv:2202.00512}, 2022.

\bibitem[Schuhmann et~al.(2022)Schuhmann, Beaumont, Vencu, Gordon, Wightman, Cherti, Coombes, Katta, Mullis, Wortsman, et~al.]{schuhmann2022laion5bopenlargescaledataset}
Christoph Schuhmann, Romain Beaumont, Richard Vencu, Cade Gordon, Ross Wightman, Mehdi Cherti, Theo Coombes, Aarush Katta, Clayton Mullis, Mitchell Wortsman, et~al.
\newblock Laion-5b: An open large-scale dataset for training next generation image-text models.
\newblock \emph{Advances in Neural Information Processing Systems}, 35:\penalty0 25278--25294, 2022.

\bibitem[Sehwag et~al.(2024)Sehwag, Kong, Li, Spranger, and Lyu]{sehwag2024stretchingdollardiffusiontraining}
Vikash Sehwag, Xianghao Kong, Jingtao Li, Michael Spranger, and Lingjuan Lyu.
\newblock Stretching each dollar: Diffusion training from scratch on a micro-budget.
\newblock \emph{arXiv preprint arXiv:2407.15811}, 2024.

\bibitem[Sohl-Dickstein et~al.(2015)Sohl-Dickstein, Weiss, Maheswaranathan, and Ganguli]{sohldickstein2015Diffusion}
Jascha Sohl-Dickstein, Eric Weiss, Niru Maheswaranathan, and Surya Ganguli.
\newblock Deep unsupervised learning using nonequilibrium thermodynamics.
\newblock In \emph{International conference on machine learning}, pages 2256--2265. PMLR, 2015.

\bibitem[Song and Ermon(2019)]{song2019generative}
Yang Song and Stefano Ermon.
\newblock Generative modeling by estimating gradients of the data distribution.
\newblock \emph{Advances in neural information processing systems}, 32, 2019.

\bibitem[Song et~al.(2020)Song, Sohl-Dickstein, Kingma, Kumar, Ermon, and Poole]{song2021scorebasedgenerativemodelingstochastic}
Yang Song, Jascha Sohl-Dickstein, Diederik~P Kingma, Abhishek Kumar, Stefano Ermon, and Ben Poole.
\newblock Score-based generative modeling through stochastic differential equations.
\newblock \emph{arXiv preprint arXiv:2011.13456}, 2020.

\bibitem[Song et~al.(2023)Song, Dhariwal, Chen, and Sutskever]{song2023consistencymodels}
Yang Song, Prafulla Dhariwal, Mark Chen, and Ilya Sutskever.
\newblock Consistency models.
\newblock \emph{arXiv preprint arXiv:2303.01469}, 2023.

\bibitem[Sun et~al.(2022)Sun, Li, and Wang]{sun2021decentralizedfederatedaveraging}
Tao Sun, Dongsheng Li, and Bao Wang.
\newblock Decentralized federated averaging.
\newblock \emph{IEEE Transactions on Pattern Analysis and Machine Intelligence}, 45\penalty0 (4):\penalty0 4289--4301, 2022.

\bibitem[Wang et~al.(2024)Wang, Zhang, Zhang, Chen, Ma, and Qu]{wang2024diffusionmodelslearnlowdimensional}
Peng Wang, Huijie Zhang, Zekai Zhang, Siyi Chen, Yi Ma, and Qing Qu.
\newblock Diffusion models learn low-dimensional distributions via subspace clustering.
\newblock \emph{arXiv preprint arXiv:2409.02426}, 2024.

\bibitem[Wortsman et~al.(2022)Wortsman, Ilharco, Gadre, Roelofs, Gontijo-Lopes, Morcos, Namkoong, Farhadi, Carmon, Kornblith, et~al.]{wortsman2022modelsoupsaveragingweights}
Mitchell Wortsman, Gabriel Ilharco, Samir~Ya Gadre, Rebecca Roelofs, Raphael Gontijo-Lopes, Ari~S Morcos, Hongseok Namkoong, Ali Farhadi, Yair Carmon, Simon Kornblith, et~al.
\newblock Model soups: averaging weights of multiple fine-tuned models improves accuracy without increasing inference time.
\newblock In \emph{International conference on machine learning}, pages 23965--23998. PMLR, 2022.

\bibitem[Yin et~al.(2024)Yin, Gharbi, Zhang, Shechtman, Durand, Freeman, and Park]{yin2024onestepdiffusiondistributionmatching}
Tianwei Yin, Micha{\"e}l Gharbi, Richard Zhang, Eli Shechtman, Fredo Durand, William~T Freeman, and Taesung Park.
\newblock One-step diffusion with distribution matching distillation.
\newblock In \emph{Proceedings of the IEEE/CVF Conference on Computer Vision and Pattern Recognition}, pages 6613--6623, 2024.

\bibitem[Yu et~al.(2024)Yu, Kwak, Jang, Jeong, Huang, Shin, and Xie]{yu2024representationalignmentgenerationtraining}
Sihyun Yu, Sangkyung Kwak, Huiwon Jang, Jongheon Jeong, Jonathan Huang, Jinwoo Shin, and Saining Xie.
\newblock Representation alignment for generation: Training diffusion transformers is easier than you think.
\newblock \emph{arXiv preprint arXiv:2410.06940}, 2024.

\bibitem[Zhao et~al.(2023)Zhao, Gu, Varma, Luo, Huang, Xu, Wright, Shojanazeri, Ott, Shleifer, et~al.]{zhao2023pytorchfsdpexperiencesscaling}
Yanli Zhao, Andrew Gu, Rohan Varma, Liang Luo, Chien-Chin Huang, Min Xu, Less Wright, Hamid Shojanazeri, Myle Ott, Sam Shleifer, et~al.
\newblock Pytorch fsdp: experiences on scaling fully sharded data parallel.
\newblock \emph{arXiv preprint arXiv:2304.11277}, 2023.

\bibitem[Zhou et~al.(2022)Zhou, Lei, Liu, Du, Huang, Zhao, Dai, Le, Laudon, et~al.]{zhou2022mixtureofexpertsexpertchoicerouting}
Yanqi Zhou, Tao Lei, Hanxiao Liu, Nan Du, Yanping Huang, Vincent Zhao, Andrew~M Dai, Quoc~V Le, James Laudon, et~al.
\newblock Mixture-of-experts with expert choice routing.
\newblock \emph{Advances in Neural Information Processing Systems}, 35:\penalty0 7103--7114, 2022.

\end{thebibliography}
}

\clearpage

\appendix
\section{Score Matching Derivation}
\label{sec:supp_math}

We provide an alternative derivation of Decentralized Flow Matching based on score matching~\cite{song2021scorebasedgenerativemodelingstochastic} rather than flow matching~\cite{lipman2023flowmatchinggenerativemodeling}.
We begin with the score, which is the gradient of the log likelihood, $p_t(x_t)$.

\begin{align}
    \nabla_{\rvx_t} \log p_t(\rvx_t)
\end{align}
By applying the chain rule, this can be expressed in terms of the derivative of the likelihood itself:
\begin{align}
    \nabla_{\rvx_t} \log p_t(\rvx_t)=  
    \frac{1}{p_t(\rvx_t)} \cdot \nabla_{\rvx_t}\; p_t(\rvx_t).
\end{align}

\noindent Let us assume our data, $x_0$, is generated in a bi-level fashion:
\begin{align}
    \rvx_0 \sim p_0(\rvx_0 | \rvv), \quad \rvv \sim p_\rvv(\rvv),
\end{align}
\noindent where $v$ is the cluster label discussed in the DDM method and $p_\rvv(\rvv)$ follows a distribution defined by the clustering procedure. The marginal likelihood \(p_t(\rvx_t)\) can then be expressed by integrating over \(\rvv\):
\begin{align}
    \nabla_{\rvx_t} \log p_t(\rvx_t)= 
    \frac{1}{p_t(\rvx_t)} \cdot
    \nabla_{\rvx_t} \sum_{\rvv} p(\rvv) \cdot
      p_t(\rvx_t |\rvv).
\end{align}

\noindent By linearity of differentiation, we distribute the gradient over the summation:
\begin{align}
    = 
    \frac{1}{p_t(\rvx_t)} \cdot
    \sum_{\rvv} p(\rvv) \cdot
    \nabla_{\rvx_t}  p_t(\rvx_t |\rvv).
\end{align}

\noindent Since the gradient of a log probability can be expressed as the probability multiplied by the gradient of its log,
\begin{align}
    = 
    \frac{1}{p_t(\rvx_t)} \cdot
    \sum_{\rvv} p(\rvv) p_t(\rvx_t | \rvv) \cdot
    \nabla_{\rvx_t}  \log p_t(\rvx_t |\rvv).
\end{align}

\noindent Finally, we invoke Bayes' Theorem:
\begin{align}
    = 
    \sum_{\rvv}  p_t(\rvv | \rvx_t) \cdot
    \nabla_{\rvx_t}  \log p_t(\rvx_t |\rvv)
\end{align}

\noindent This result mirrors the flow matching derivation, showing that the score prediction for the overall data distribution can be recast as a linear combination of score predictions for each data cluster. Each learned expert predicts its own conditional score, which ensemble according to the posterior probability of the expert label given the latent $\rvx_t$. 

\section{Additional Training and Evaluation Details}

Our split of LAION Aesthetics contains 153.6 million image-caption pairs. We pretrain all diffusion models on 256x256 square crop images encoded through Huggingface's finetuned Stable Diffusion VAE (sd-vae-ft-mse). This encoder employs an 8× spatial downsampling factor. Throughout training, we maintain a patch size of 2 for best quality, resulting in a pretraining context length of 256.

For high-resolution finetuning, we choose five aspect ratio buckets to handle varying image dimensions while maintaining consistent tokenized sequence length (3600). The bucket are as follows:

\begin{itemize}
    \item \textbf{1280 × 720} (16:9 landscape)
    \item \textbf{1200 × 768} (~3:2 landscape)
    \item \textbf{960 × 960} (square)
    \item \textbf{768 × 1200} (~2:3 portrait)
    \item \textbf{720 × 1280} (9:16 portrait)
\end{itemize}

Images are mapped to their nearest matching bucket by aspect ratio. Following best practices from Stable Diffusion 3~\cite{esser2024scalingrectifiedflowtransformers}, we adjust the timestep schedule for high-resolution training and inference by applying a log-SNR shift of 3~\cite{hoogeboom2023simplediffusionendtoenddiffusion}. We also modify the Rotary Positional Embedding (RoPE) in the MMDiT architecture by interpolating RoPE inputs within the central square region and extrapolating for peripheral areas.

For evaluation, we use standard classifier-free guidance scales: 7.5 for LAION text-conditional generation and 3 for ImageNet class-conditional generation. All evaluations use 50 sampling steps to ensure consistent comparisons. We compute FID, CLIP-FID and DINO-FID metrics on fixed evaluation splits to standardize evaluations.

\begin{figure*}[ht!]
    \centering
    \begin{subfigure}[t]{0.44\linewidth}
        \centering
        \includegraphics[width=\linewidth]{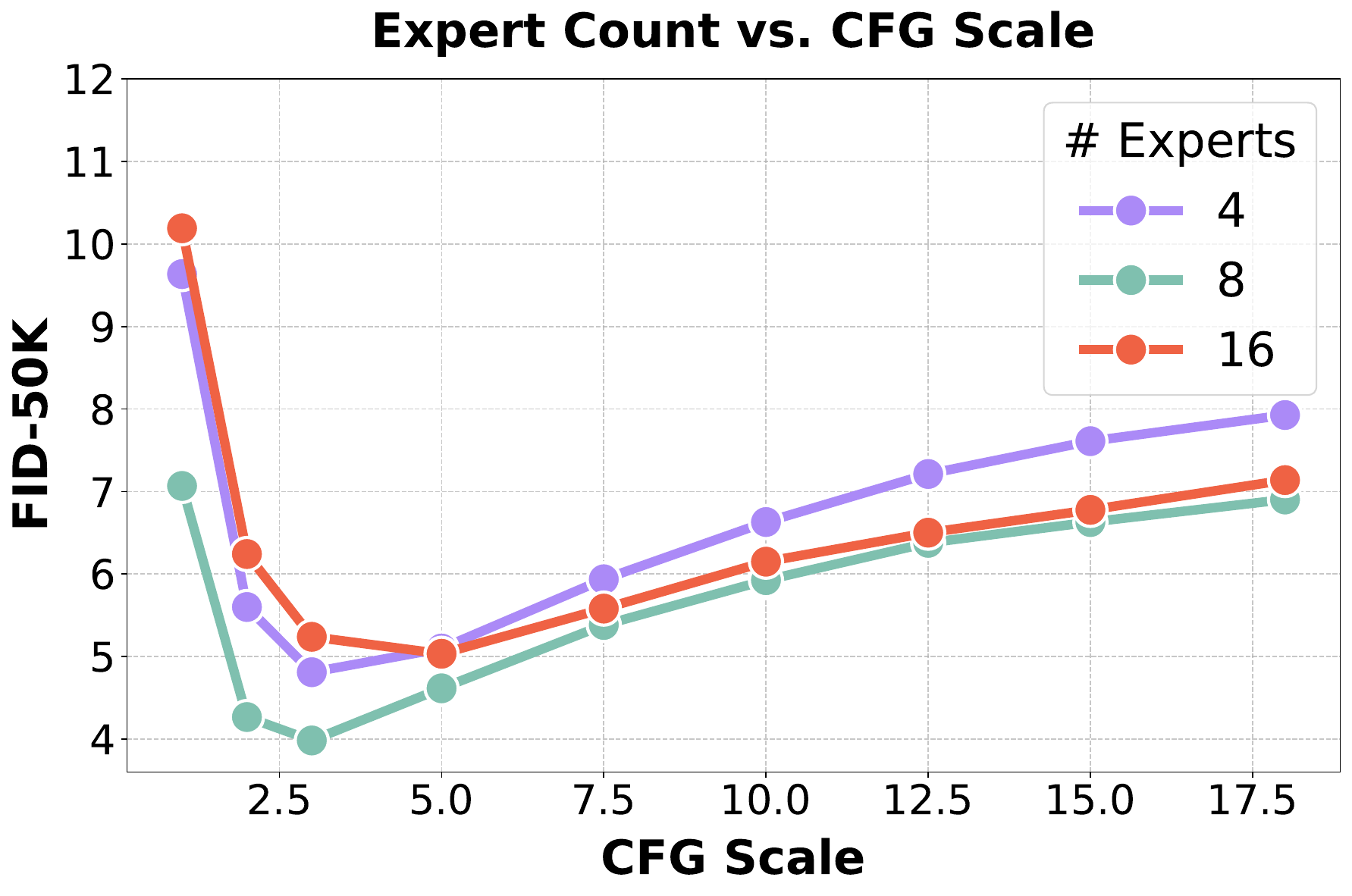}
          \vspace{-1.5em}
        \caption{}
        \label{fig:laion_cfg_expert_count}
    \end{subfigure}
    \hspace{20pt}
    \begin{subfigure}[t]{0.44\linewidth}
        \centering
        \includegraphics[width=\linewidth]{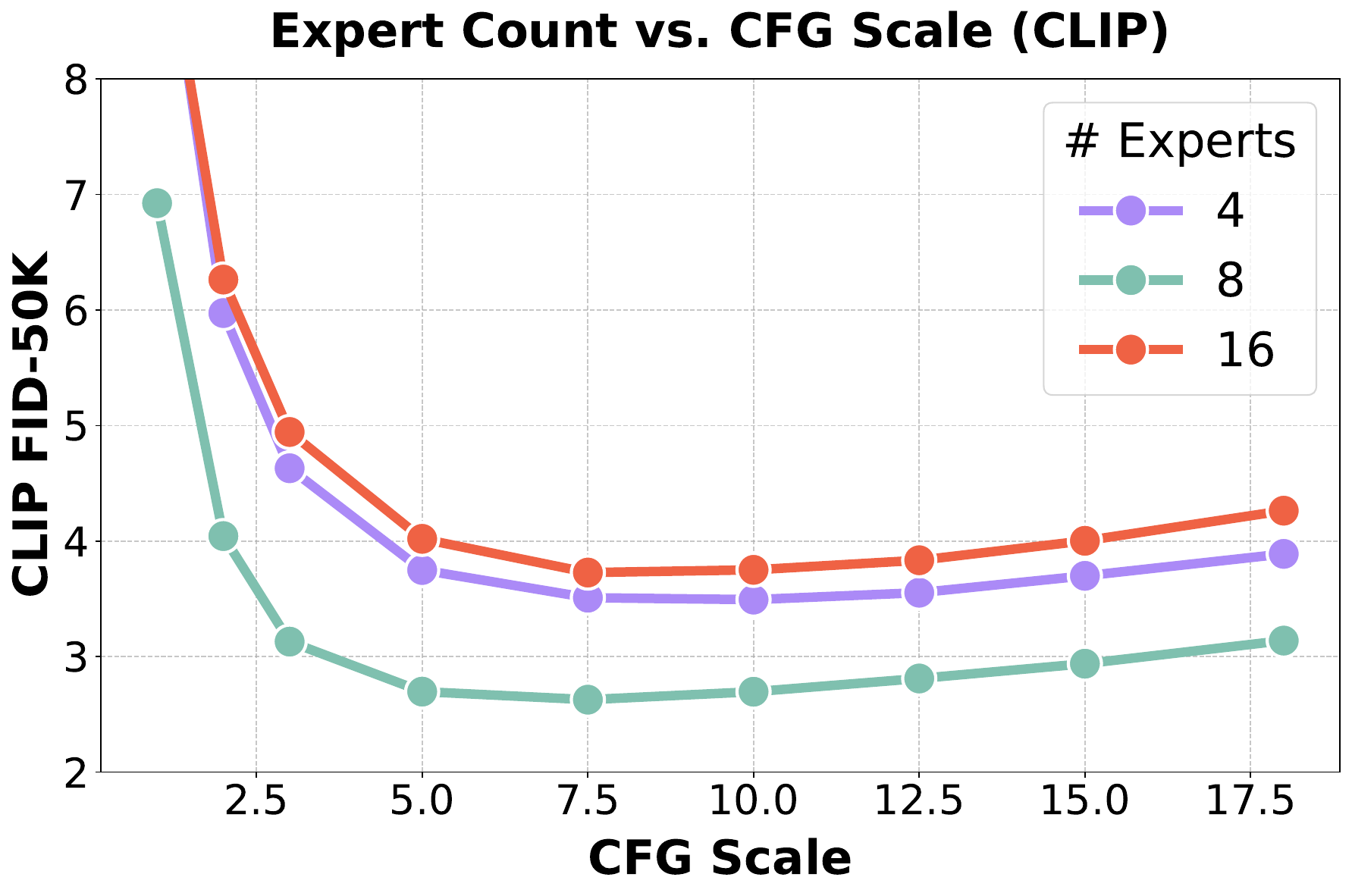}
          \vspace{-1.5em}
        \caption{}
        \label{fig:laion_clip_cfg_expert_count}
    \end{subfigure}

    \begin{subfigure}[t]{0.44\linewidth}
        \centering
        \includegraphics[width=\linewidth]{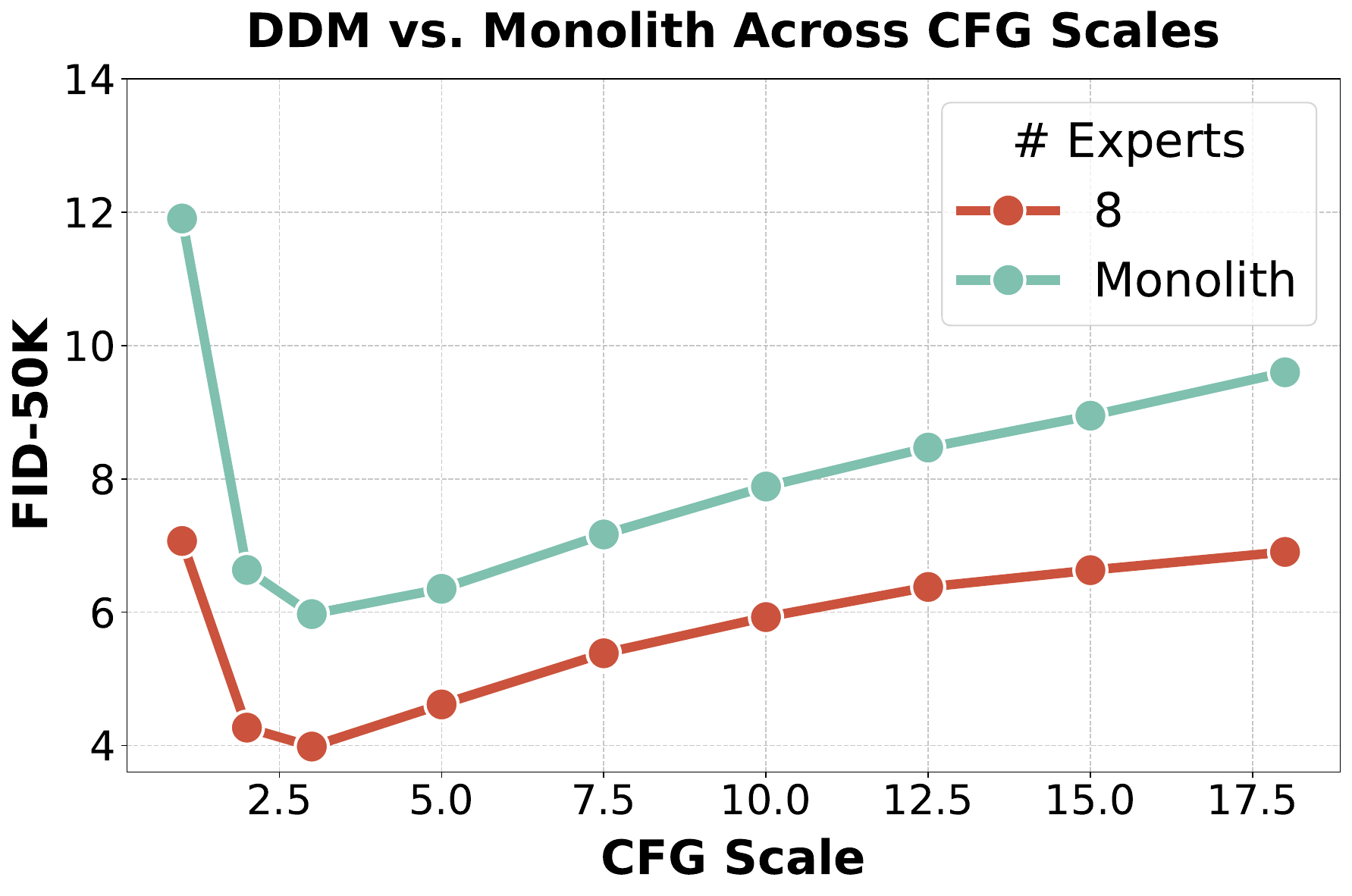}
          \vspace{-1.5em}
        \caption{}
        \label{fig:laion_cfg_8_experts_vs_monolith}
    \end{subfigure}
    \hspace{20pt}
    \begin{subfigure}[t]{0.44\linewidth}
        \centering
        \includegraphics[width=\linewidth]{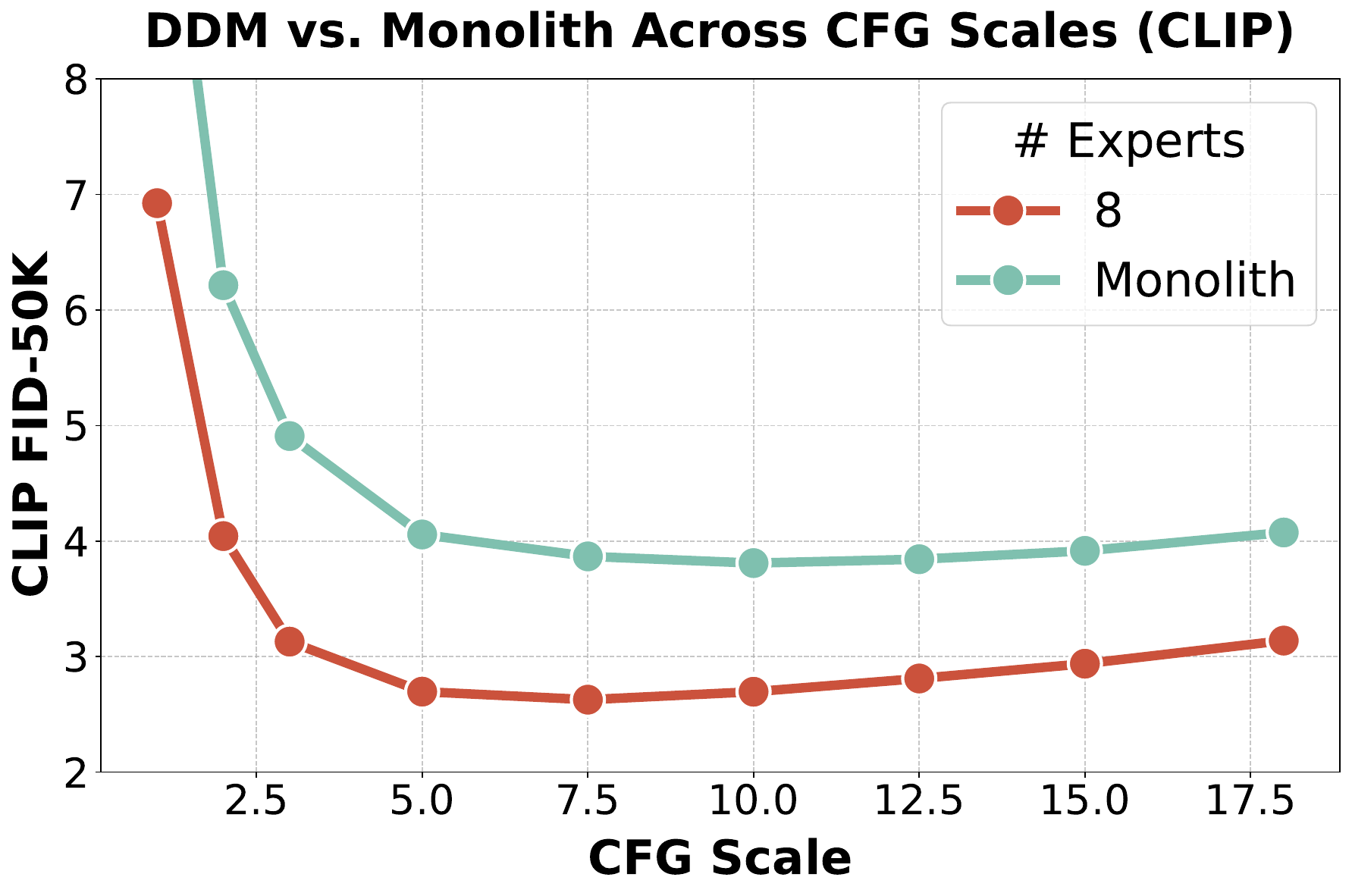}
          \vspace{-1.5em}
        \caption{}
        \label{fig:laion_clip_cfg_8_experts_vs_monolith}
    \end{subfigure}
    \begin{subfigure}[t]{0.44\linewidth}
        \centering
        \includegraphics[width=\linewidth]{figures/distillation/laion_distillation.pdf}
        \caption{}
        \label{fig:laion_distillation_fid}
\end{subfigure}
\hspace{20pt}
\begin{subfigure}[t]{0.44\linewidth}
        \centering
        \includegraphics[width=\linewidth]{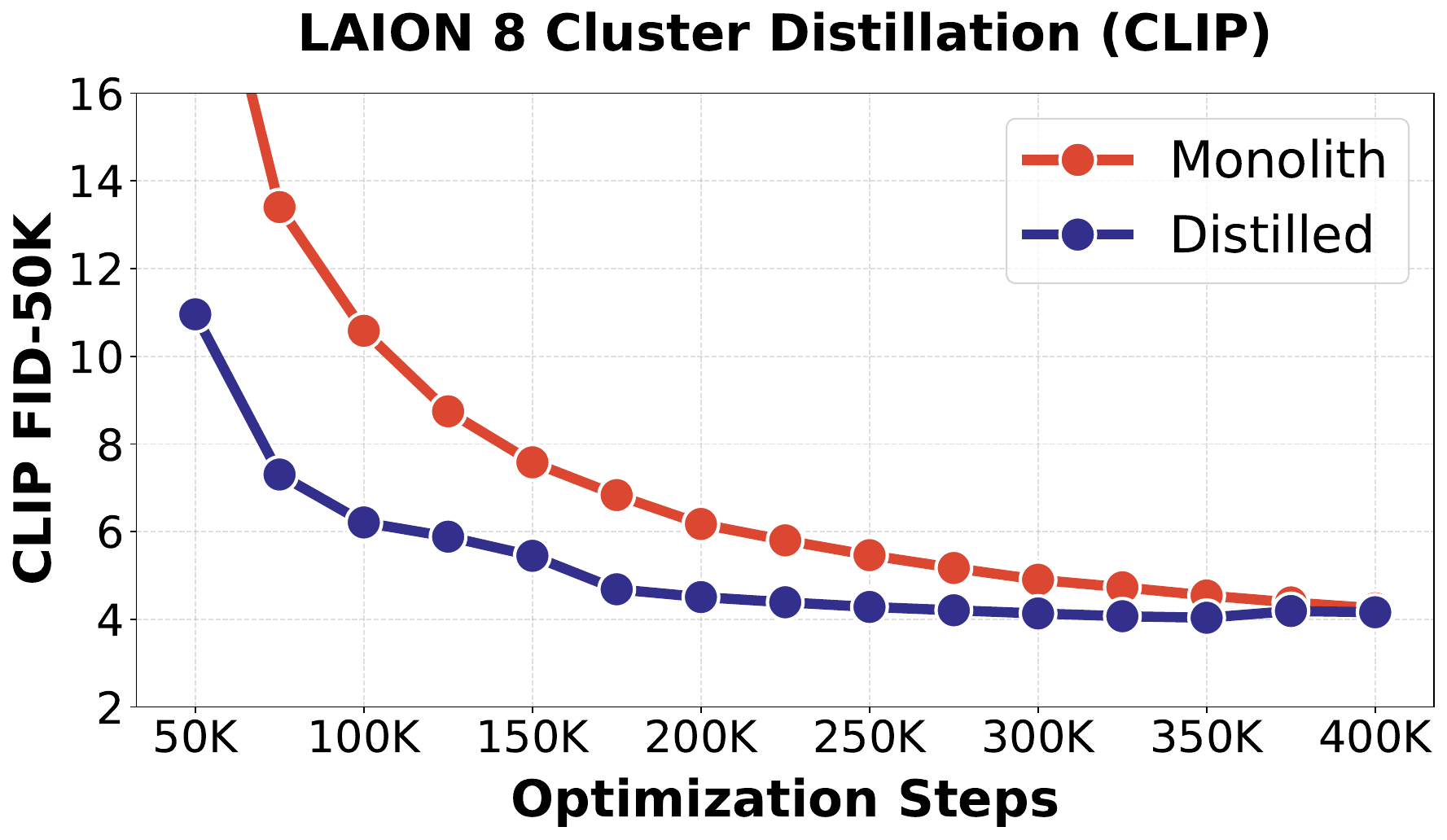}
        \caption{}
        \label{fig:laion_distillation_clip}
\end{subfigure}
    \vspace{-0.6em}
    \caption{\textbf{Additional Quantitative Analysis.} We sweep CFG scales across decentralized and monolith diffusion models trained on LAION Aesthetics (a, b, c, d), finding that optimal CFG scales are consistent across models. Distillation matches the performance of training a monolith from raw data at a fraction of the FLOP-cost (e, f).}
    \label{fig:supp_plots}
    \vspace{-1.4em}
\end{figure*}

\section{Additional Quantitative Analysis}

Our additional quantitative analysis explores key test-time DDM hyperparameters. We test various ensemble combinations in Table~\ref{tab:supp_inference_ablations}, including nucleus sampling which is common in LLM decoding. Top-1 sampling consistently delivers the best performance while being the most computationally efficient. This finding holds across different expert counts, router temperatures, and threshold probabilities.

Our classifier-free guidance (CFG)~\cite{ho2022classifierfreediffusionguidance} scale experiments (Figure~\ref{fig:supp_plots}) show that DDMs respond similarly to monolith models, suggesting that standard CFG scales can be directly applied to DDMs. Additionally, our training efficiency analysis (Figures~\ref{fig:laion_distillation_fid} and~\ref{fig:laion_distillation_clip}) confirms that distillation achieves comparable generation quality (FID) with only one-quarter of the monolith's batch size (256 vs 1024).

\section{Additional Qualitative Results}

\begin{figure*}[ht!]
    \centering
    \includegraphics[width=0.99\linewidth]{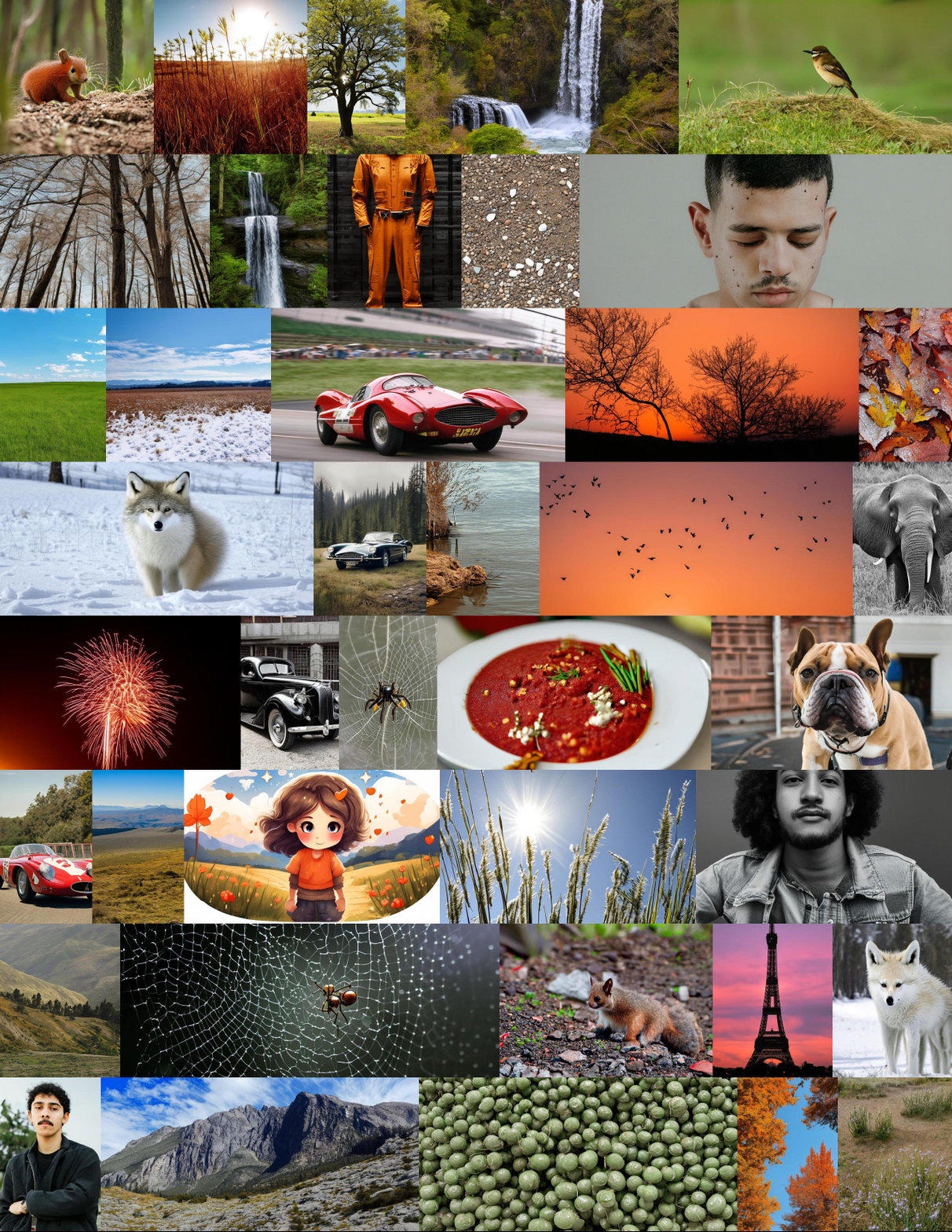}
    \caption{\textbf{Additional Selected Samples.}
}
    \label{fig:additonal_selected}
\end{figure*}

\begin{figure*}[ht!]
    \centering
    \includegraphics[width=0.99\linewidth]{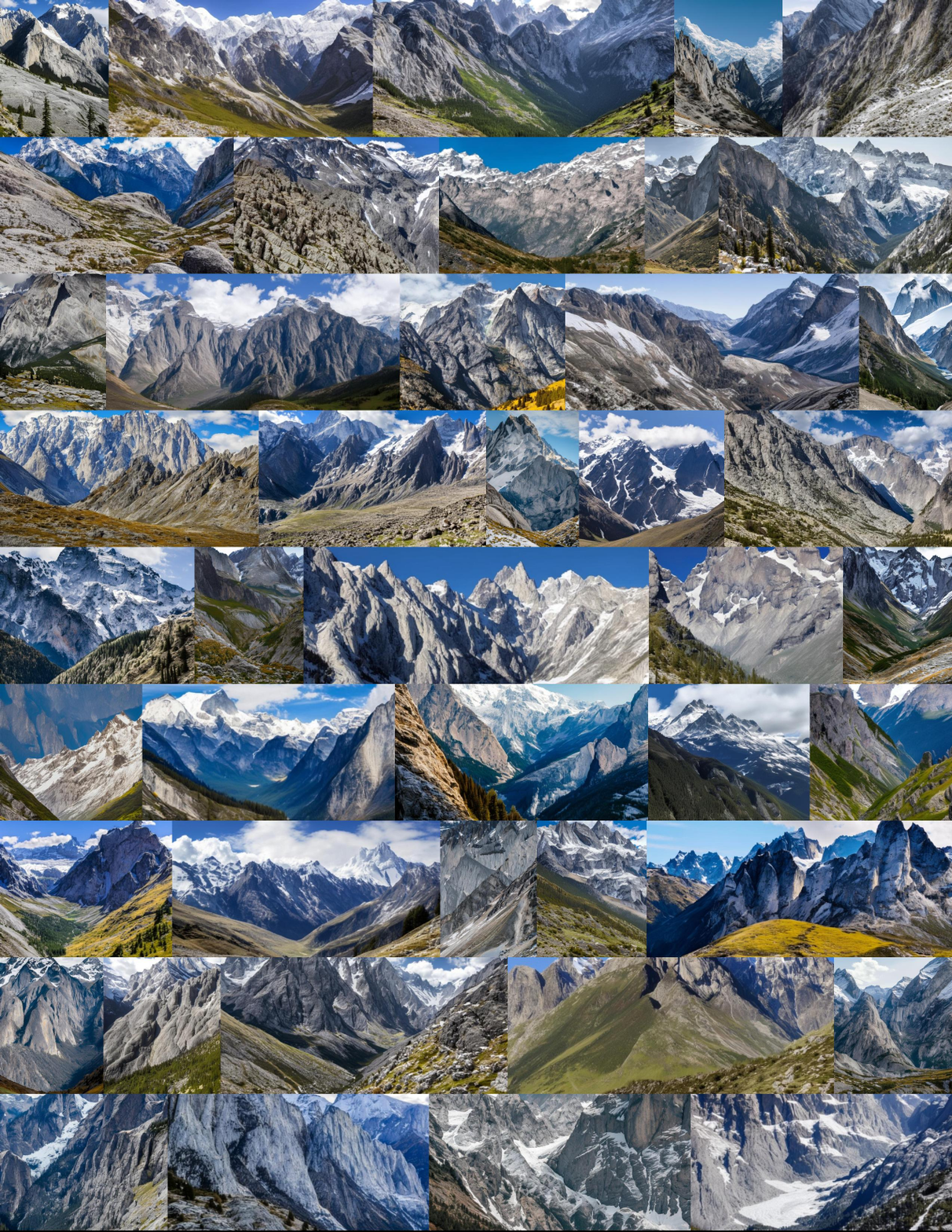}
    \caption{\textbf{Random Samples, Fixed Prompt.}
    a photo of the dolomites
}
    \label{fig:random_samples_dolomites}
\end{figure*}

\begin{figure*}[ht!]
    \centering
    \includegraphics[width=0.99\linewidth]{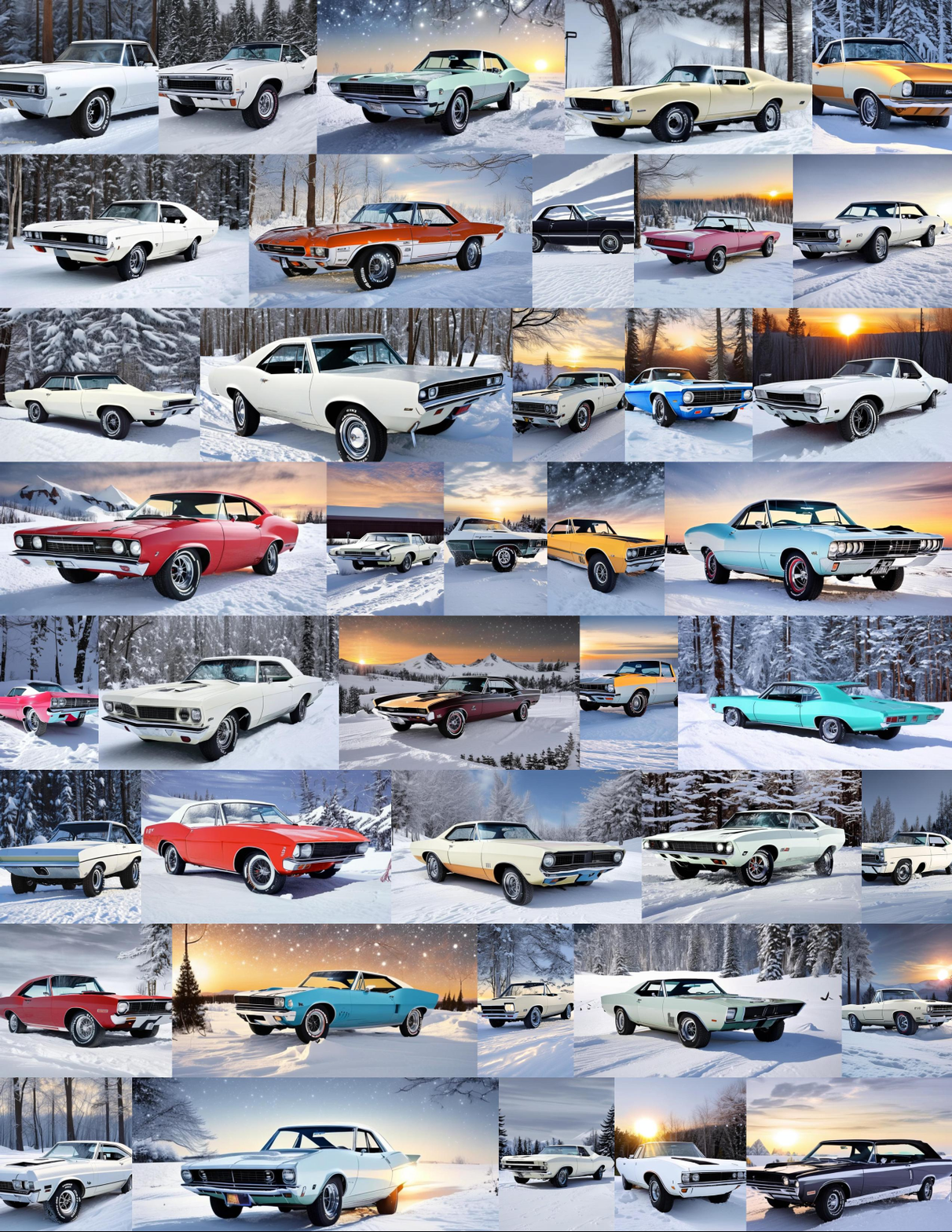}
    \caption{\textbf{Random Samples, Fixed Prompt.}
    1969 Polaris Colt, restored to showroom, static display in snow, winter sunrise
}
    \label{fig:random_samples_polaris}
\end{figure*}

\begin{figure*}[ht!]
    \centering
    \includegraphics[width=0.99\linewidth]{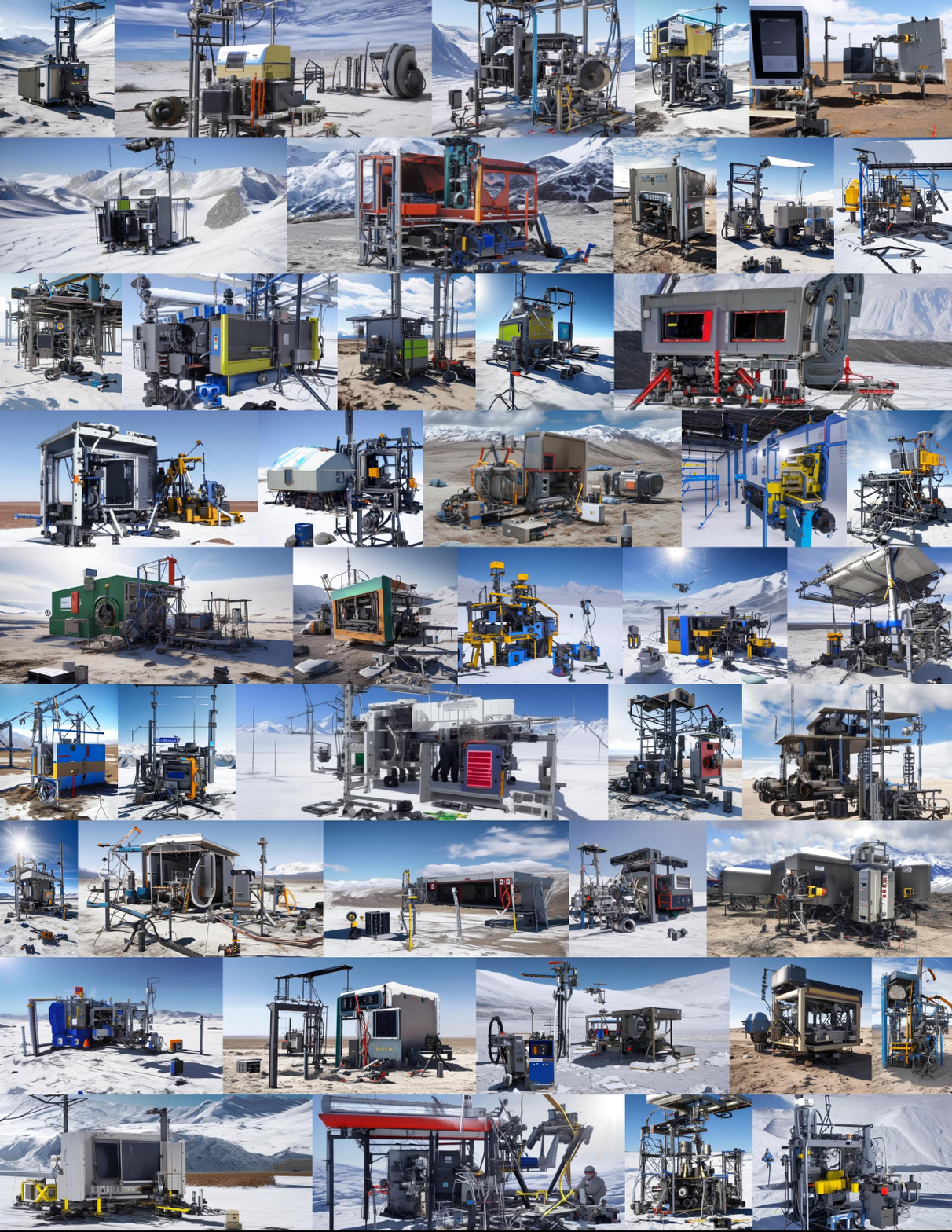}
    \caption{\textbf{Random Samples, Fixed Prompt.}
    weather research station in extreme conditions, monitoring equipment, natural elements
}
    \label{fig:random_samples_weather_station}
\end{figure*}

\begin{figure*}[ht!]
    \centering
    \includegraphics[width=0.99\linewidth]{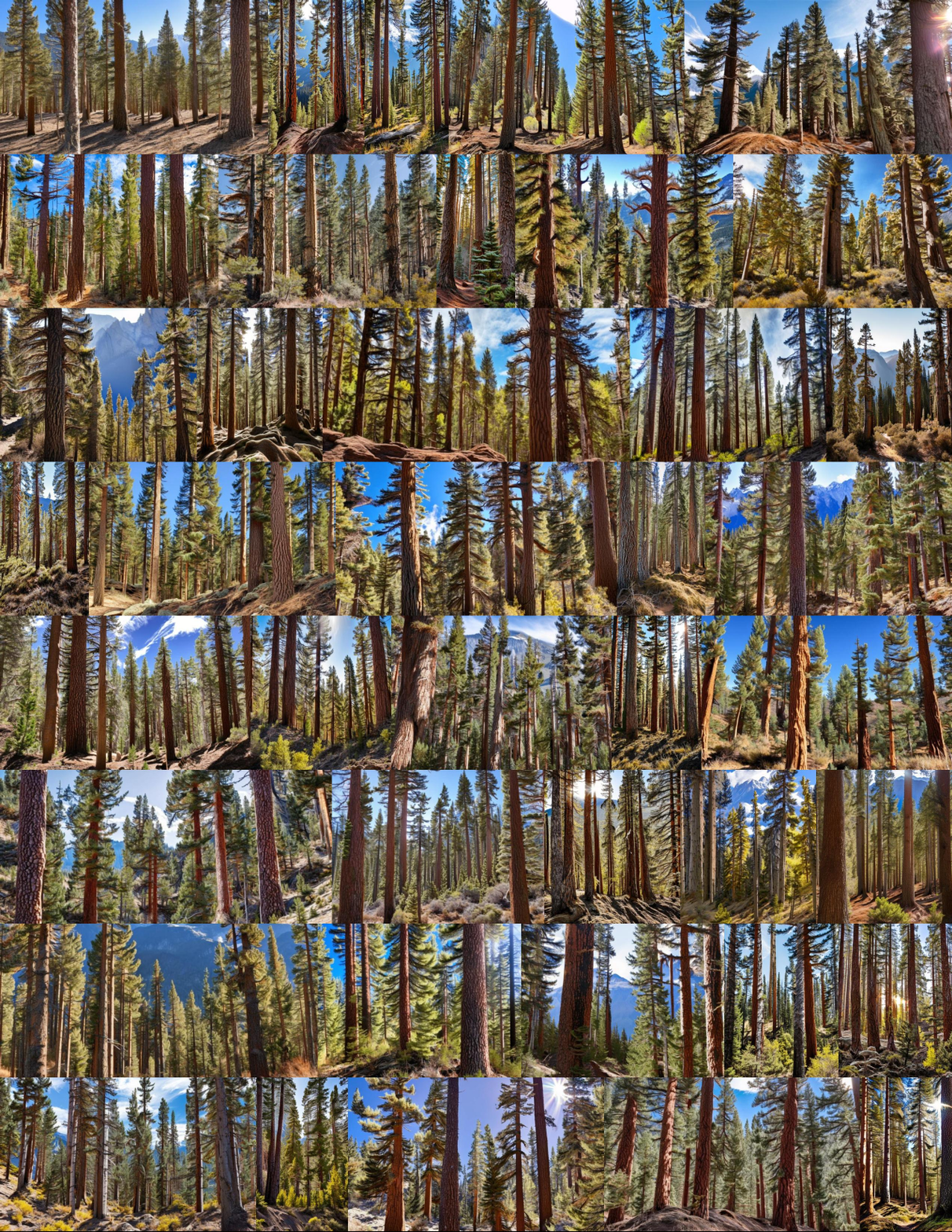}
    \caption{\textbf{Random Samples, Fixed Prompt.}
    ancient bristlecone pine forest, twisted trees, high-altitude light, rugged mountain backdrop
}
    \label{fig:random_samples_bristlecone}
\end{figure*}

\begin{figure*}[ht!]
    \centering
    \includegraphics[width=0.99\linewidth]{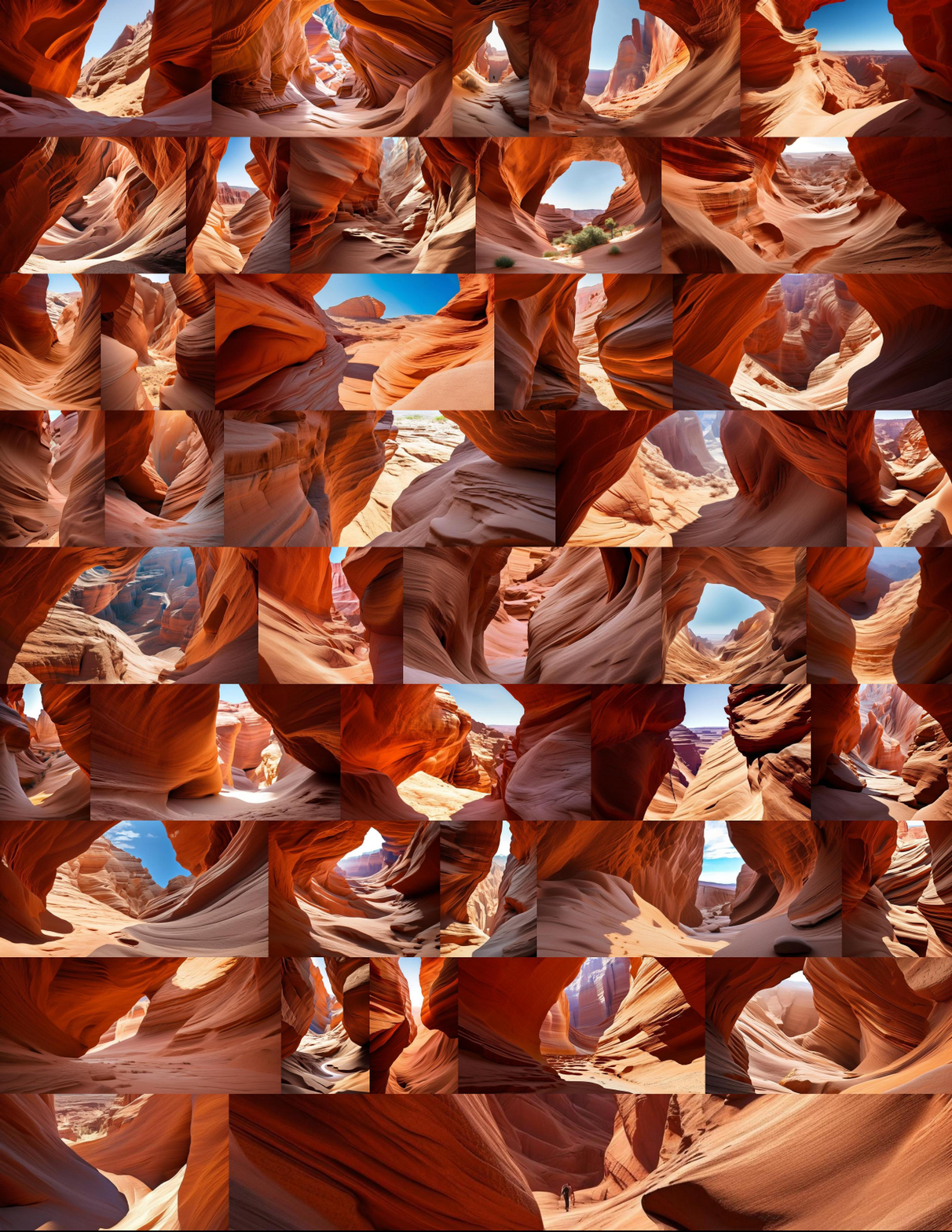}
    \caption{\textbf{Random Samples, Fixed Prompt.}
    deep desert slot canyon, sandstone textures, shaft of light, natural color gradients
}
    \label{fig:random_samples_canyon}
\end{figure*}

\begin{figure*}[ht!]
    \centering
    \includegraphics[width=0.99\linewidth]{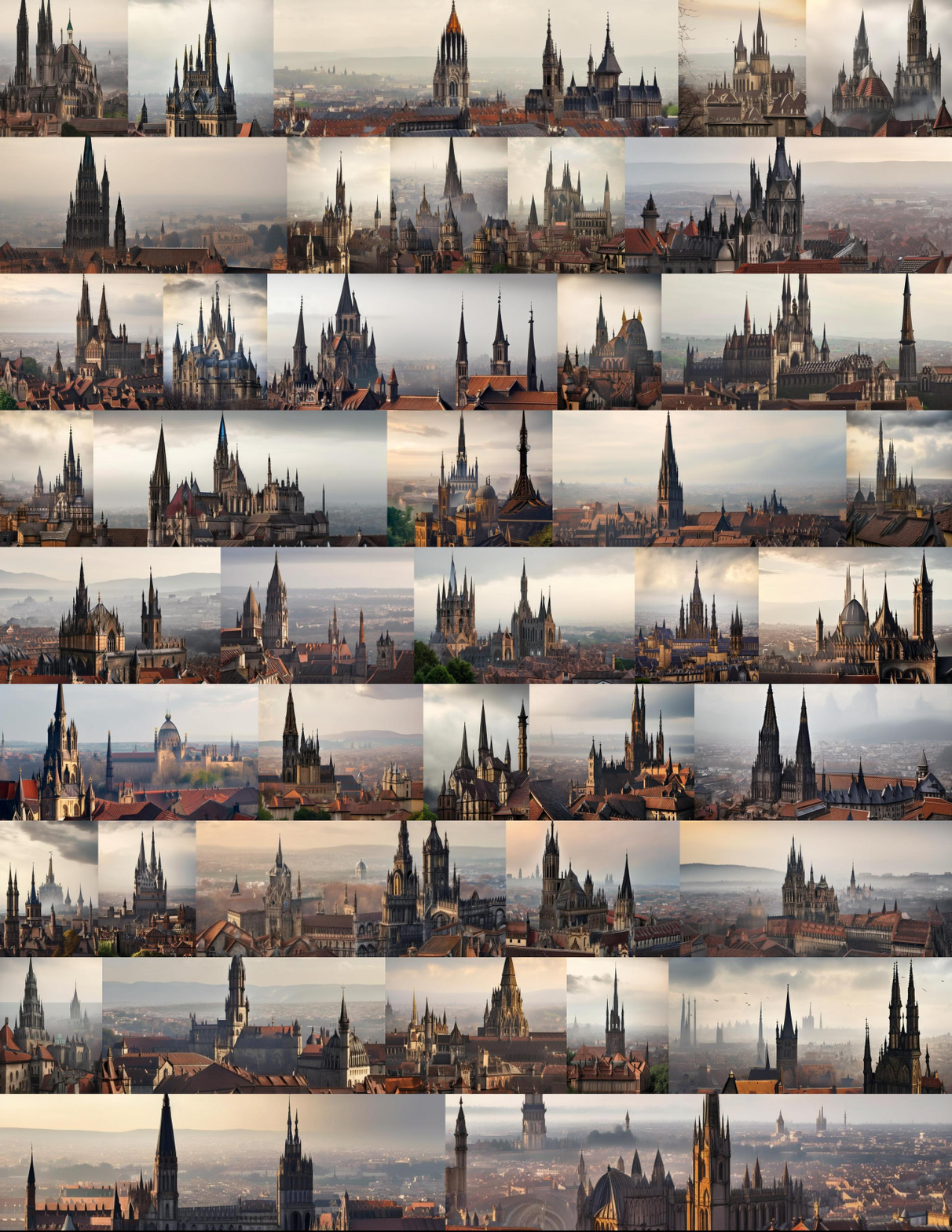}
    \caption{\textbf{Random Samples, Fixed Prompt.}
    gothic cathedral spires piercing morning mist, ancient European city roofscape
}
    \label{fig:random_samples_cathedral}
\end{figure*}

\begin{figure*}[ht!]
    \centering
    \includegraphics[width=0.99\linewidth]{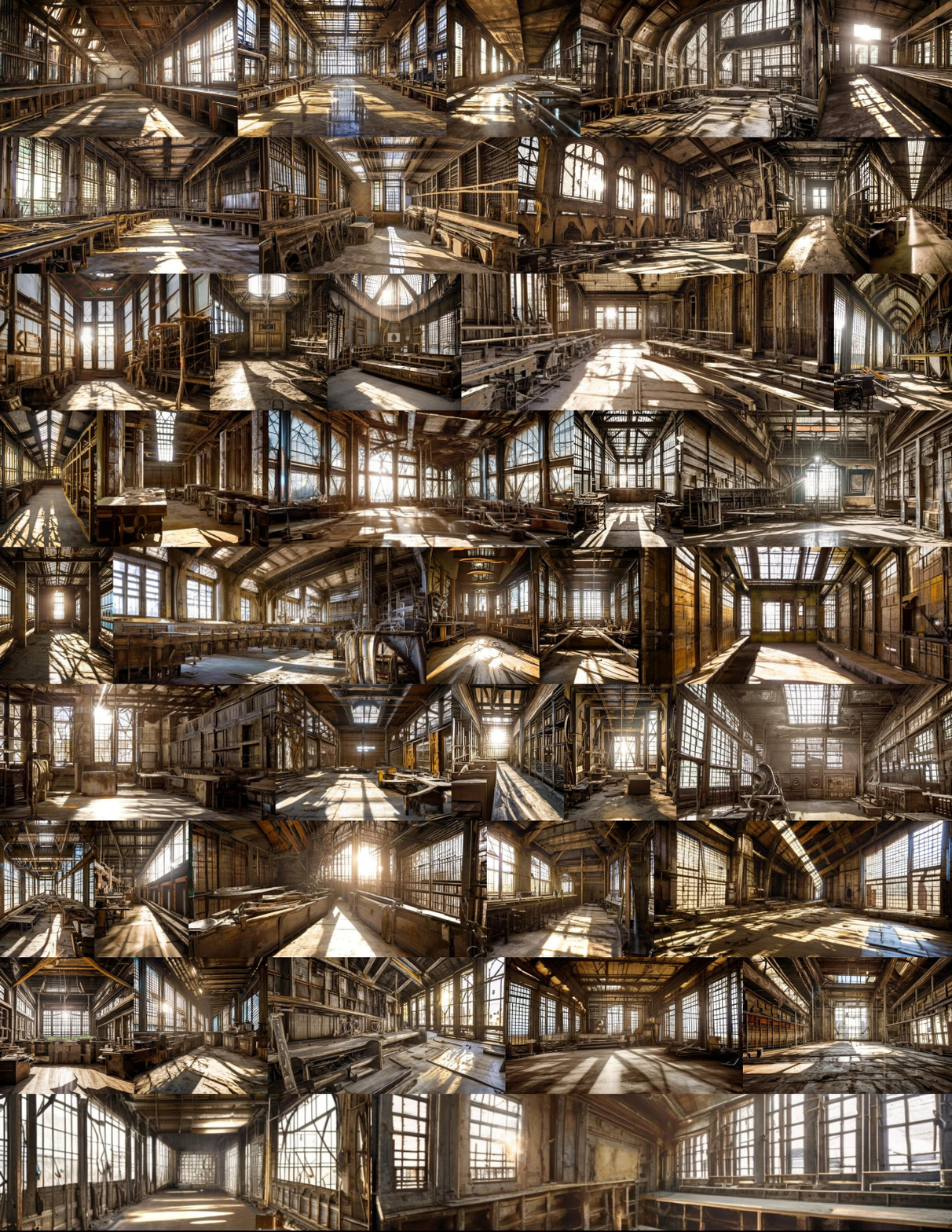}
    \caption{\textbf{Random Samples, Fixed Prompt.}
    historic textile mill interior, preserved machinery, sunbeams through industrial windows
}
    \label{fig:random_samples_textile_mill}
\end{figure*}

\begin{figure*}[ht!]
    \centering
    \includegraphics[width=0.99\linewidth]{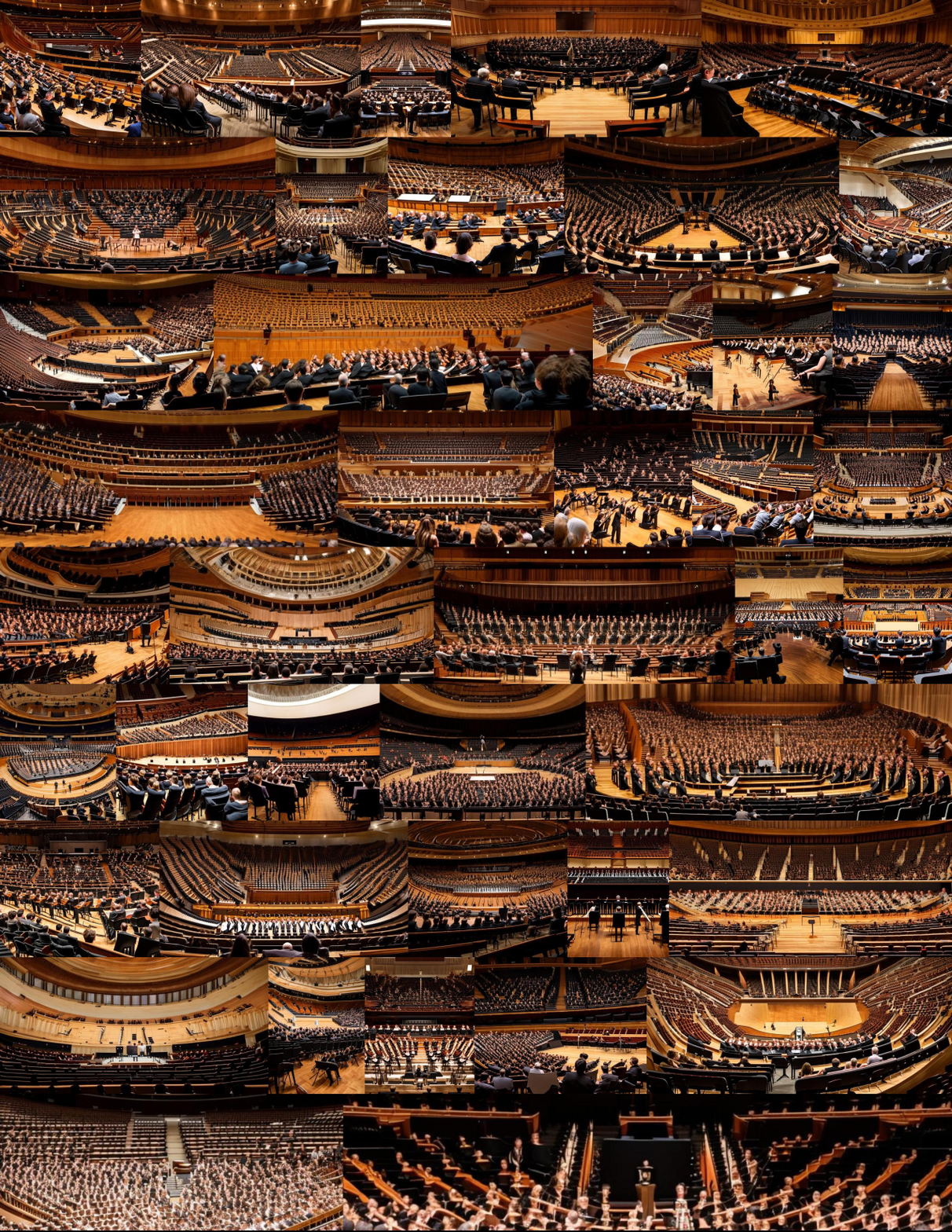}
    \caption{\textbf{Random Samples, Fixed Prompt.}
    symphony orchestra during rehearsal, conductor's perspective, historic concert hall
}
    \label{fig:random_samples_symphony}
\end{figure*}

We provide additional selected samples from our largest DDM ensemble in Figure~\ref{fig:additonal_selected} as well as random samples for different text prompts in Figures~\ref{fig:random_samples_dolomites} through~\ref{fig:random_samples_symphony}.

\begin{table*}[!t]
\begin{center}

  \small
\begin{tabular}{@{}p{1.6cm}p{0.8cm}cp{1.1cm}p{0.3cm}cp{2.2cm}p{2.2cm}p{2.2cm}@{}}
\toprule
{Inference Strategy} & {Expert Count} & {Temp.} & {Active Experts} & {$p$} & {GFLOPs $\downarrow$} & {FID $\downarrow$} & {CLIP FID $\downarrow$} & {DINO FID $\downarrow$} \\ \midrule
Monolith  & 1  & -   & 1 & 0.00 & 308  & 12.81  & 5.58  & 343.96  \\
\midrule
Full      & 4  & -   & 4 & 0.00 & 1245 & 12.75  & 6.82  & 386.3   \\
Top-1     & 4  & -   & 1 & 0.00 & \textbf{334}  & \textbf{12.54}  & 6.72  & \textbf{378.4}   \\
Top-2     & 4  & -   & 2 & 0.00 & 642  & 12.76  & 6.75  & 384.12  \\
Top-3     & 4  & -   & 3 & 0.00 & 950  & 12.88  & 6.8   & 385.79  \\
Sample    & 4  & 0.5 & 1 & 0.00 & \textbf{334}  & 117.02 & 36.25 & 1321.17 \\
Sample    & 4  & 1.0 & 1 & 0.00 & \textbf{334}  & 89.14  & 28.01 & 1042.09 \\
Sample    & 4  & 2.0 & 1 & 0.00 & \textbf{334}  & 15.08  & 7.53  & 425.29  \\
Sample    & 4  & 0.5 & 2 & 0.00 & 642  & 12.67  & \textbf{6.71}  & 380.85  \\
Sample    & 4  & 1.0 & 2 & 0.00 & 642  & 12.67  & 6.73  & 382.93  \\
Sample    & 4  & 2.0 & 2 & 0.00 & 642  & 13.51  & 7.05  & 400.08  \\
Sample    & 4  & 0.5 & 3 & 0.00 & 950  & 12.57  & 6.76  & 381.44  \\
Sample    & 4  & 1.0 & 3 & 0.00 & 950  & 12.84  & 6.81  & 385.18  \\
Sample    & 4  & 2.0 & 3 & 0.00 & 950  & 13.59  & 7.08  & 400.23  \\
Nucleus   & 4  & 0.5 & 1 & 0.90 & \textbf{334}  & 15.74  & 9.99  & 412.49  \\
Nucleus   & 4  & 1.0 & 1 & 0.90 & \textbf{334}  & 15.72  & 10.04 & 411.31  \\
Nucleus   & 4  & 2.0 & 1 & 0.90 & \textbf{334}  & 17.35  & 10.31 & 432.46  \\
Threshold & 4  & 1.0 & - & 0.01 & -    & 12.82  & 6.81  & 385.44  \\
Threshold & 4  & 1.0 & - & 0.05 & -    & 12.67  & 6.73  & 382.54  \\
Threshold & 4  & 1.0 & - & 0.10 & -    & 12.63  & 6.76  & 382.86  \\
\midrule
Full      & 8  & -   & 4 & 0.00 & 2490 & 10.52  & 5.85  & 354.15  \\
Top-1     & 8  & -   & 1 & 0.00 & \textbf{334}  & \textbf{9.85}   & \textbf{5.54}  & \textbf{339.56}  \\
Top-2     & 8  & -   & 2 & 0.00 & 642  & 10.33  & 5.73  & 349.28  \\
Top-3     & 8  & -   & 3 & 0.00 & 950  & 10.45  & 5.77  & 351.91  \\
Sample    & 8  & 1.0 & 1 & 0.00 & \textbf{334}  & 190.95 & 59.03 & 2105.79 \\
Sample    & 8  & 2.0 & 1 & 0.00 & \textbf{334}  & 184.06 & 50.55 & 1790.24 \\
Sample    & 8  & 0.5 & 2 & 0.00 & 642  & 9.93   & 5.57  & 343.51  \\
Sample    & 8  & 1.0 & 2 & 0.00 & 642  & 10.28  & 5.72  & 348.39  \\
Sample    & 8  & 2.0 & 2 & 0.00 & 642  & 17.11  & 8.09  & 471.18  \\
Sample    & 8  & 0.5 & 3 & 0.00 & 950  & 10.04  & 5.62  & 342.86  \\
Sample    & 8  & 1.0 & 3 & 0.00 & 950  & 10.42  & 5.78  & 350.91  \\
Sample    & 8  & 2.0 & 3 & 0.00 & 950  & 12.06  & 6.38  & 380.54  \\
Nucleus   & 8  & 0.5 & 1 & 0.90 & \textbf{334}  & 188.66 & 60.09 & 2110.22 \\
Nucleus   & 8  & 1.0 & 1 & 0.90 & \textbf{334}  & 152.16 & 48.37 & 1609.23 \\
Nucleus   & 8  & 2.0 & 1 & 0.90 & \textbf{334}  & 33.9   & 14    & 682.31  \\
Threshold & 8  & 1.0 & - & 0.01 & -    & 10.51  & 5.82  & 351.17  \\
Threshold & 8  & 1.0 & - & 0.05 & -    & 10.32  & 5.73  & 349.86  \\
Threshold & 8  & 1.0 & - & 0.10 & -    & 10.18  & 5.7   & 346.9   \\
\midrule
Full      & 16 & -   & 4 & 0.00 & 4980 & 15.43  & 7.57  & 440.54  \\
Top-1     & 16 & -   & 1 & 0.00 & \textbf{334}  & \textbf{12.51}  & \textbf{6.6}   & \textbf{397.99}  \\
Top-2     & 16 & -   & 2 & 0.00 & 642  & 148.26 & 41.13 & 1535.85 \\
Top-3     & 16 & -   & 3 & 0.00 & 950  & 91.76  & 29.53 & 1105.92 \\
Sample    & 16 & 1.0 & 1 & 0.00 & \textbf{334}  & 232.1  & 71.88 & 2557.41 \\
Sample    & 16 & 2.0 & 1 & 0.00 & \textbf{334}  & 259    & 81.49 & 2797.76 \\
Sample    & 16 & 0.5 & 2 & 0.00 & 642  & 161.29 & 47.76 & 1732.88 \\
Sample    & 16 & 1.0 & 2 & 0.00 & 642  & 174.23 & 54.41 & 1941.49 \\
Sample    & 16 & 0.5 & 3 & 0.00 & 950  & 119.84 & 40.18 & 1510.78 \\
Sample    & 16 & 1.0 & 3 & 0.00 & 950  & 44.62  & 20.04 & 772.02  \\
Sample    & 16 & 2.0 & 3 & 0.00 & 950  & 26.01  & 10.92 & 603.31  \\
Threshold & 16 & 1.0 & - & 0.01 & -    & 14.92  & 7.44  & 431.89  \\
Threshold & 16 & 1.0 & - & 0.05 & -    & 12.62  & 6.61  & 399.09  \\
Threshold & 16 & 1.0 & - & 0.10 & -    & 12.69  & 6.61  & 398.33 \\
 \bottomrule
 \label{supp_table}
\end{tabular}

\vspace{-0.5em}
    \caption{\textbf{Test-Time Combination Strategies.} We ablate strategies and relevant hyperparameters for sampling from our ImageNet DDM ensemble at test-time. Across many experiments, we find that simply selecting the top expert outperforms more sophisticated alternatives.}
    \vspace{-2em}
  \label{tab:supp_inference_ablations}%
      
\end{center}
\end{table*}

\clearpage

\end{document}